\pdfoutput=1

\documentclass[11pt]{article}

\usepackage[]{acl}

\usepackage{times}
\usepackage{latexsym}
\usepackage{CJK}

\usepackage[T1]{fontenc}

\usepackage[utf8]{inputenc}

\usepackage{microtype}

\usepackage{inconsolata}

\newcommand{\ignore}[1]{}
\usepackage{multirow}
\usepackage{graphicx} 
\usepackage{enumitem}

\title{Benchmarking Large Language Models on CFLUE - \\A Chinese Financial Language Understanding Evaluation Dataset}

\author{Jie Zhu$^1$, Junhui Li$^2$\thanks{Corresponding Author}, Yalong Wen$^{1}$, \textbf{Lifan Guo}$^1$\\
$^1$Alibaba Group, Hangzhou, China\\
$^2$School of Computer Science and Technology, Soochow University, Suzhou, China\\
{zhujie951121@gmail.com},
{lijunhui@suda.edu.cn}\\
{\{wenyalong.wyl, lifan.lg\}@alibaba-inc.com}
}

\begin{document}
\begin{CJK}{UTF8}{gkai}
\maketitle
\begin{abstract}
In light of recent breakthroughs in large language models (LLMs) that have revolutionized natural language processing (NLP), there is an urgent need for new benchmarks to keep pace with the fast development of LLMs. In this paper, we propose CFLUE, the Chinese Financial Language Understanding Evaluation benchmark, designed to assess the capability of LLMs across various dimensions. Specifically, CFLUE provides datasets tailored for both knowledge assessment and application assessment. In knowledge assessment, it consists of 38K+ multiple-choice questions with associated solution explanations. These questions serve dual purposes: answer prediction and question reasoning. In application assessment, CFLUE features 16K+ test instances across distinct groups of NLP tasks such as text classification, machine translation, relation extraction, reading comprehension, and text generation. Upon CFLUE, we conduct a thorough evaluation of representative LLMs. The results reveal that only GPT-4 and GPT-4-turbo achieve an accuracy exceeding 60\% in answer prediction for knowledge assessment, suggesting that there is still substantial room for improvement in current LLMs. In application assessment, although GPT-4 and GPT-4-turbo are the top two performers, their considerable advantage over lightweight LLMs is noticeably diminished. The datasets and scripts associated with CFLUE are openly accessible at \url{https://github.com/aliyun/cflue}. 

\end{abstract}

\section{Introduction}

Recently, the remarkable capabilities exhibited by large language models (LLMs) have brought significant advancements and revolutionized natural language processing (NLP). In response to the emergence of LLMs, new benchmarks have been introduced to assess a varied range of abilities exhibited by these models. Notable examples include MMLU~\cite{hendrycks_etal_iclr_2021_mmlu}, HELM~\cite{liang_etal_2023_tmlr_helm}, BIG-bench~\cite{srivastava_etal_2023_tmlr_big}, and GLUE~\cite{wang_etal_iclr_2019_glue}, which are widely employed to evaluate LLMs' capabilities in English. Additionally, benchmarks such as CLEU~\cite{xu_etal_2020_coling_clue}, CMMLU~\cite{li_etal_arxiv_2023_cmmlu}, and CEval~\cite{huang_etal_2023_nips_ceval} are widely used to evaluate their capabilities in Chinese. 

However, assessing LLMs in the financial domain, especially within the Chinese context, presents considerable challenges due to limitations in existing Chinese financial evaluation datasets. Notably, datasets like FinanceIQ~\cite{duxiaoman_2022_financeiq} and FinEval~\cite{zhang_etal_arxiv_2023_fineval} have restrictions in terms of size and diversity, hindering a comprehensive evaluation of LLM performance. While FinanceIQ and FinEval focus solely on multiple-choice tasks, there is a lack of datasets allowing a thorough examination of LLM capabilities in text generation. Additionally, existing shared tasks, such as those in CCKS~\cite{tianchi_2019_ccks_entity,tianchi_2020_ccks_entity,tianchi_2021_ccks_causal,tianchi_2022_ccks}, predominantly concentrate on event extraction tasks, limiting the objective and quantitative measurement of LLM performance.

Inspired by the FLUE benchmark~\cite{shah_etal_emnlp_2019_flue}, which encompasses a comprehensive set of datasets across five financial domain tasks in English, this paper introduces a novel dataset named CFLUE (Chinese Financial Language Understanding Evaluation). CFLUE addresses the aforementioned challenges by providing a benchmark for evaluating LLM performance through various NLP tasks, categorized into knowledge assessment and application assessment. In the knowledge assessment section, CFLUE contains 38K+ questions from 15 types of financial qualification mock exams of varying difficulty levels and subjects. These questions are in the multiple-choice question format to enable standardized and objective evaluations. Additionally, each question is accompanied by a solution explanation, facilitating open-ended or chain-of-thought assessments of LLMs' reasoning capabilities. In the application assessment section, CFLUE contains 16K+ test instances covering five groups of classical NLP tasks: text classification, machine translation, relation extraction, reading comprehension, and text generation. These instances are sourced from existing shared tasks or are annotated by professionals using real data sources.

Based on CFLUE, we assess the effectiveness of several representative LLMs across general and financial domains. This includes an examination of three OpenAI LLMs, nine lightweight LLMs in the general domain, and three lightweight LLMs in the financial domain. From the experimental results, we have the following key findings:

\begin{itemize}[leftmargin=*]
\item The findings are not consistent in terms of tasks in knowledge assessment and application assessment. In knowledge assessment, GPT-4 and GPT-4-turbo~\cite{openai_arxiv_2023_gpt4} demonstrate significant superiority over other LLMs, achieving over 60\% accuracy in answer prediction. However, this suggests substantial room for improvement in current LLMs. In application assessment, although GPT-4 and GPT-4-turbo remain top performers on average, and in specific tasks, they may even fall behind certain lightweight LLMs. This could be attributed to the specialized design of these lightweight LLMs, which are tailored for Chinese data. 
\item Existing financial domain LLMs, such as FinGPT V3~\cite{yang_2023_ijcai_fingpt}, DISC-FinLLM~\cite{chen_etal_2023_arxiv_finllm}, and Tongyi-Finance~\cite{tongyi_2023_tongyifinance}, exhibit poor zero-shot performance in both knowledge and application assessment tasks, indicating limited coverage of financial knowledge and significant room for improvement.
\item Lightweight LLMs benefit significantly from supervised fine-tuning. For instance, ChatGLM3-6B~\cite{zeng_etal_arxiv_2022_glm}, Qwen-7B~\cite{bai_etal_2023_arxiv_qwen}, and Baichuan2-7B~\cite{baichuan_2023_baichuan2} achieve comparable or superior performance to ChatGPT~\cite{openai_2022_chatgpt} in both answer prediction and reasoning tasks, despite having only 4\% of ChatGPT's parameters.
\end{itemize}

In summary, this paper provides valuable insights into the performance of LLMs in Chinese financial contexts from multiple perspectives. Our findings suggest that there still exists much room for improvement even for the current best performers. We hope that CFLUE could guide the developers to understand the abilities of their models from multiple dimensions and facilitate the growth of foundation models in Chinese financial domain.

\section{Related Work}

\begin{table*}[htb]
\centering
\small
\begin{tabular}{l|l|l}
\hline
\bf Language & \bf Source & \bf Question Type (Task) \\
\hline
\multirow{9}{*}{English} & FINQA~\cite{chen_etal_emnlp_2021_finqa} & Question Answering \\
\cline{2-3}
& TAT-QA~\cite{zhu_etal_acl_2021_tat} & Question Answering \\
\cline{2-3}
& BizBench~\cite{kedziorski_arxiv_2023_bizbench} & Quantitative Reasoning \\
\cline{2-3}
& FINANCEBENCH~\cite{islam_arxiv_2023_financebench} & Question Answering \\
\cline{2-3}
& \multirow{6}{*}{FLUE~\cite{shah_etal_emnlp_2019_flue}} & Sentiment Classification \cite{malo_etal_2014_jasist_fpb} \\
& & Sentiment Analysis (FiQA~\cite{fiqa_2018_fiqa})\\
& & News Headline Classification~\cite{sinha_and_khandait_2020_arxiv} \\
& & Named Entity Recognition~\cite{alvarado_etal_alta_2015} \\
& & Structure Boundary Detection~\cite{finsbd3_2021_finsbd3} \\
& &  Question Answering (FiQA~\cite{fiqa_2018_fiqa}) \\
\hline
\multirow{7}{*}{Chinese} & CCKS~\cite{tianchi_2019_ccks_entity,tianchi_2020_ccks_entity,tianchi_2021_ccks_causal,tianchi_2022_ccks} & Event Extraction, Event Entity (and Causality) Extraction\\
\cline{2-3}
& FinanceIQ~\cite{duxiaoman_2022_financeiq} & Multiple-choice Question Answering\\
\cline{2-3}
& FinEval~\cite{zhang_etal_arxiv_2023_fineval} & Multiple-choice Question Answering \\
\cline{2-3}
& FinGPT-fineval~\cite{yang_2023_ijcai_fingpt} & Multiple-choice Question Answering\\
\cline{2-3}
& \multirow{5}{*}{CFLUE (Ours)} & Multiple-choice Question Answering \& Reasoning \\
& & Text Classification \\
& & Machine Translation \\
& & Relation Extraction \\
& & Reading comprehension\\
& & Text Generation \\
\hline
\end{tabular}
\caption{A summary of benchmarks in the financial domain.}
\label{tab:related_datasets}
\end{table*}

\noindent\textbf{Financial Evaluation Datasets.} A summary of recent benchmarks in the financial domain is presented in Table~\ref{tab:related_datasets}. In English, FINQA by \citet{chen_etal_emnlp_2021_finqa} introduces a dataset comprising 8,281 question-answering pairs, emphasizing numerical reasoning processes. Both TAT-QA by \citet{zhu_etal_acl_2021_tat} build a large-scale QA dataset containing both Tabular And Textual data from real financial reports. BizBench by \citet{kedziorski_arxiv_2023_bizbench} consists of 8 quantitative reasoning tasks of question-answering (QA) for structured and unstructured financial data via program synthesis. FINANCEBENCH by~\citet{islam_arxiv_2023_financebench} consists of 10,231 questions about publicly traded companies, with corresponding answers and evidence strings. FLUE by \citet{shah_etal_emnlp_2019_flue} utilizes datasets from existing literature to establish comprehensive benchmark across five tasks in financial domain, including sentiment analysis classification from Financial PhraseBank dataset~\cite{malo_etal_2014_jasist_fpb} and sentiment analysis regression from FiQA 2018 Task-1~\citet{maia_etal_www_2018},\footnote{\url{https://sites.google.com/view/fiqa}} named entity recognition (NER) from NER data on loan agreement~\cite{alvarado_etal_alta_2015}, question answering from FiQA 2018 Shared Task-2~\citet{maia_etal_www_2018}, news headline classification from gold commodity news~\cite{sinha_and_khandait_2020_arxiv}, and structure boundary detection from FinSBD-3 Shared Task~\cite{finsbd3_2021_finsbd3}. 

In Chinese, the CCKS series since 2019 CCKS has released various datasets for event extraction tasks~\cite{tianchi_2019_ccks_entity,tianchi_2020_ccks_entity,tianchi_2021_ccks_causal,tianchi_2022_ccks}. Additionally, both FinanceIQ and FinEval offer thousands of multiple-choice question answering pairs that could be used as evaluation suites for LLMs~\cite{duxiaoman_2022_financeiq,zhang_etal_arxiv_2023_fineval}. However, compared to English, Chinese datasets tend to focus on either event extraction~\cite{tianchi_2022_ccks} or multiple-choice question-answering \cite{duxiaoman_2022_financeiq,zhang_etal_arxiv_2023_fineval,yang_2023_ijcai_fingpt}, demonstrating limited task diversity. In contrast, our paper introduces CFLUE (Chinese Financial Language Understanding Evaluation), featuring a set of heterogeneous benchmark tasks for a more comprehensive evaluation. Additionally, beyond evaluation suites, there are financial datasets like SmoothNLP\footnote{\url{https://github.com/smoothnlp/FinancialDatasets}}, IREE~\cite{ren_etal_ccks_2022}, suitable for training or fine-tuning models in the finance domain. 

\noindent\textbf{Other Benchmark Datasets.} The development of LMs~\cite{devlin_etal_2019_naacl_bert,radford_etal_2019_gpt} has witnessed heterogeneous benchmarks to probe their diverse abilities. In English, conventional benchmarks traditionally target single (type) tasks such as natural language understanding~\cite{wang_etal_iclr_2019_glue,wang_etal_nips_2019_superglue}, reading comprehension~\cite{rajpurkar_etal_acl_2018,dua_etal_2019_naacl_drop}, and reasoning~\cite{zellers_etal_2019_acl_hellaswag,sakaguchi_etal_2020_acm_winogrande}, etc. Recently researchers introduce various benchmarks with broader knowledge or domain expertise to comprehensively assess the capabilities of LLMs. For example, TruthfulQA~\cite{lin_etal_2022_acl_truthfulqa} evaluates the truthfulness of language models across 38 categories, spanning health, law, finance and politics. MMLU~\cite{hendrycks_etal_iclr_2021_mmlu} provides a multi-task evaluation that covers 57 tasks in the fields like elementary mathematics, US history, computer science, and law.
BIG-bench~\cite{srivastava_etal_2023_tmlr_big} and HELM~\cite{liang_etal_2023_tmlr_helm} benchmarks cover as much to 204 and 42 tasks, respectively. 

In Chinese, CLEU~\cite{xu_etal_2020_coling_clue} is the first large-scale Chinese benchmark that covers nive tasks such as single-sentence/sentence-pair classification, reading comprehension, etc. AIGEval~\cite{zhong_etal_arxiv_2023_agieval} aims to assess the capabilities of LMs in human-centric standardized exams, including college entrance exams, law school admission tests, math competitions, and lawyer qualification tests. Both CMMLU~\cite{li_etal_arxiv_2023_cmmlu} and CEval~\cite{huang_etal_2023_nips_ceval} are comprehensive Chinese benchmarks spanning multiple levels and diverse subjects. CMExam~\cite{liu_etal_2023_nips_cmexam} focus on the Chinese medical domain, providing multiple-choice questions from the Chinese National Medical Licensing Examination.

\section{CFLUE: Chinese Financial Language Understanding Evaluation}

\begin{figure*}[t]
\centering
\includegraphics[width=6.5in, trim={0cm 0cm 0cm 0cm}, clip]{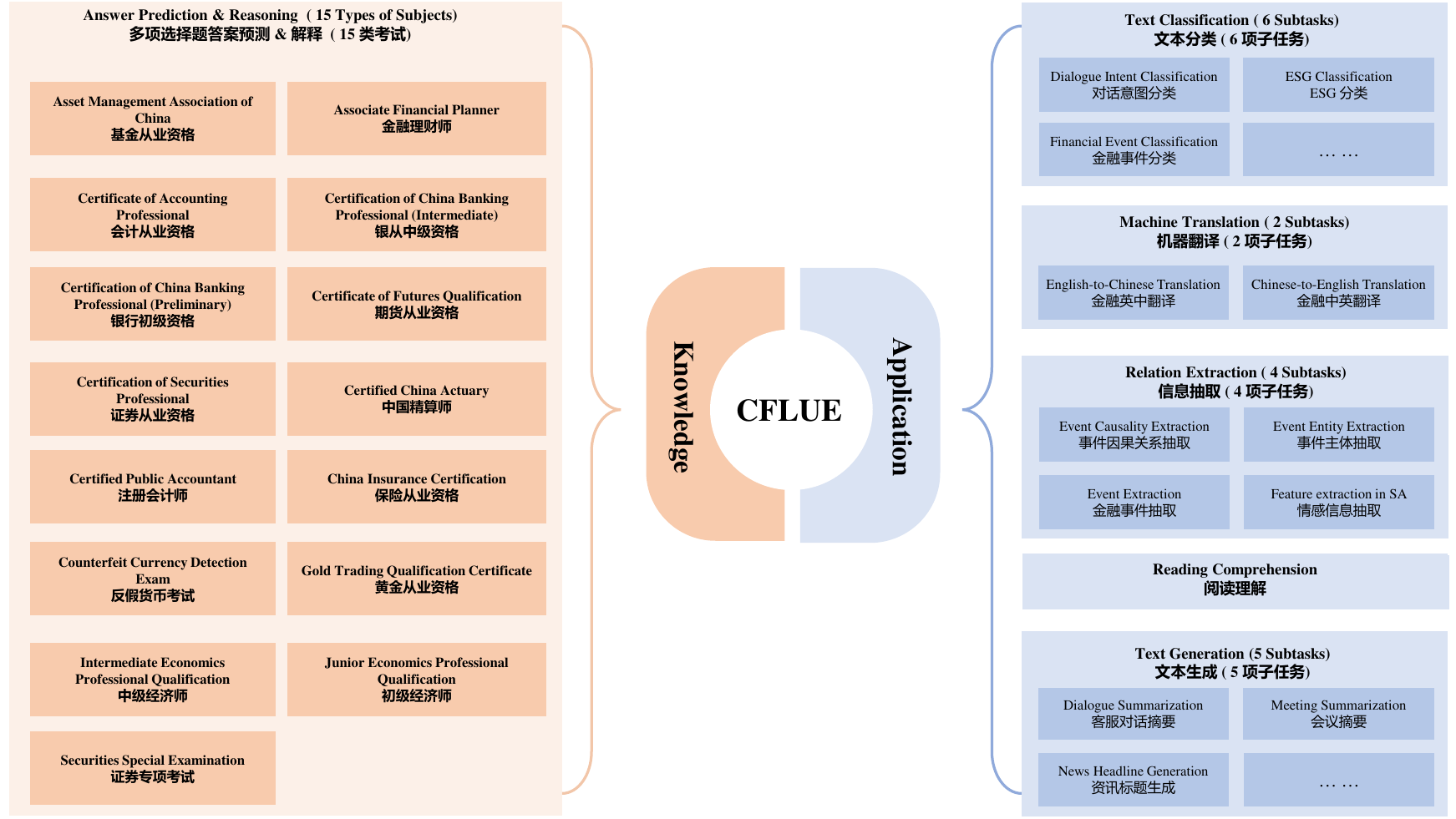}
\caption{Overview diagram of CFLUE benchmark.}
\label{fig:overview}
\end{figure*}

\begin{table*}
\centering
\small
\begin{tabular}{l|l|lll|l|l}
\hline
\multirow{2}{*}{\bf Type} & \multirow{2}{*}{\bf Task} & \multicolumn{3}{c|}{\bf Size} & \multirow{2}{*}{\bf Avg. Length} & \multirow{2}{*}{\bf Description}\\
& & \bf Train & \bf Valid & \bf Test & & \\
\hline
Knowl. & Answer Prediction \& Reasoning & 30,908 & 3,864 & 3,864 & 130 & 15 types of qualification exams\\
\hline
\multirow{6}{*}{Appl.} & Text Classification & - & - & 3,312 & 840 & 6 subtasks\\
 & Machine Translation & - & - & 3,000 & 51 & 2 translation directions\\
 & Relation Extraction & - & - & 3,500 & 274 & 4 subtasks\\
 & Reading Comprehension & - & - & 2,710 & 201 & in question answering format\\
 & Text Generation & - & - & 4,000 & 947 & 5 subtasks\\
\cline{2-7}
 & Total & - & - & 16,522 & - & - \\
\hline
\end{tabular}
\caption{Statistics of CFLUE. The average length indicates the average number of words in the inputs.}
\label{tab:statistics}
\end{table*}

\subsection{Tasks and Data Collection}
While various LLMs may exhibit similar performance on tasks within the general domain, it is often the domain-specific tasks that set them apart. The primary objective of CFLUE is to assess the financial domain knowledge capabilities of models from multiple dimensions. As illustrated in Figure~\ref{fig:overview}, CFLUE evaluates the proficiency of LLMs from both knowledge and application perspective. 

\noindent\paragraph{Knowledge assessment:} This section consists of questions presented in a multiple-choice format, sourced from real-world, challenging mock exams accessible through public channels. The primary origins include 15 types of qualification exams covering diverse difficulty levels and subjects. 60\% of the questions feature up to six choices, with only one being the correct answer. Additionally, 10\% of the questions are true-false, while the remainder includes options where more than one choice may be correct. The utilization of multiple-choice questions in this assessment section ensures a fair and objective evaluation the performance of LLMs. Moreover, each question is associated with its solution explanation which could be used to evaluate the reasoning capability of LLMs. As shown in Figure~\ref{fig:multi-choice} of Appendix~\ref{sec:data_knowledge_assessment}, each question has a brief solution explanation annotated by professionals. For detailed data collection of knowledge assessment, please refer to Appendix~\ref{sec:data_knowledge_assessment}.

\noindent\paragraph{Application assessment:} This section consists of five sets of real-word financial NLP tasks: 

\begin{itemize}[leftmargin=*]
\item Text classification, which consists of six subtasks, covering dialogue intent classification, ESG (environment, social and governance) classification, event classification, etc. Each subtask features an individual taxonomy, ranging from 3 to 77 classes. Most instances are collected from financial online websites and annotated by professionals. 

\item Machine translation, which consists of Chinese-to-English (Zh$\rightarrow$En) and English-to-Chinese (En$\rightarrow$Zh) translation subtasks. Both subtasks share a dataset collected from bilingual economic news reports.\footnote{\url{https://www.kekenet.com/menu/13650/}}

\item Relation extraction, which consists of four subtasks, namely event extraction, feature extraction in sentiment analysis, event causality extraction, and event entity extraction. Specifically, test instances are randomly selected from DUEE-fin~\cite{han_etal_2022_nlpcc_duee} for event extraction, from CCKS 2021~\cite{tianchi_2021_ccks_causal} for event causal relation extraction, and from CCKS 2019~\cite{tianchi_2019_ccks_entity} for event entity extraction. Instances for feature extraction in sentiment analysis are collected from real data of a financial firm and are annotated by professionals.

\item Reading comprehension, which is in the form of question answering. It takes a passage (or document) of a text and and a corresponding question as inputs and gives their answer as output. We first crawl news reports, insurance product documents, research reports from financial websites and construct a database with 90K documents. Then we invite financial experts to generate questions. For a question $q$, we use a house-build retriever\ignore{\footnote{It is built upon \url{https://www.langchain.com/} and \url{https://huggingface.co/moka-ai/m3e-base}.}} to retrieve the most relevant document $d$ from the database. Therefore, $d$ may or not contain answer to question $q$. If it contains, annotators label the answer from $d$; otherwise, they mark the answer as \textit{根据提供的背景信息，无法回答该问题。(According to the provided information, I cannot answer that question)}.

\item Text generation, which consists of five subtasks, namely dialogue summarization, meeting summarization, news headline generation, research report headline generation and term interpretation. For meeting summarization, 500 audio files of meetings from brokerage firms are collected and transformed into texts via ASR toolkit.\footnote{\url{https://github.com/alibaba-damo-academy/FunASR}} Professionals manually annotate the summarizations. For dialogue summarization, we randomly select test instances from CSDS~\cite{lin_etal_emnlp_2021_csds}. For news headline generation and research report headline generation, we simply collect news and research reports from EastMoney website.\footnote{\url{https://www.eastmoney.com/}} Finally, financial terminologies and their explanations are extracted from ChinaValue website.\footnote{\url{http://www.chinavalue.net/Wiki}} 
\end{itemize}

Appendix~\ref{sec:data_application_assessment} presents more details for data collection of application assessment and quality control.

\noindent\textbf{Efforts to Mitigate Data Contamination:} To mitigate data contamination, we adopt strategies outlined in \citet{huang_etal_2023_nips_ceval}. The mock exams utilized in the knowledge assessment are exclusively sourced from PDF or Microsoft Word documents available on the Internet, avoiding direct extraction from plain text webpages. We carefully extract multiple-choice questions, their answers and solution explanations using pdfplumber\footnote{ \url{https://github.com/jsvine/pdfplumber}} and PaddleOCR\footnote{ \url{https://github.com/PaddlePaddle/PaddleOCR}}, followed by manual verification of the final structured format. In an additional step to minimize the risk of data contamination, we employ GPT-4 to rephrase the questions and shuffle the multiple choices. The test instances in the application assessment can be categorized into three groups based on their data sources. Firstly, some instances are extracted from datasets in shared tasks, typically released in zipped files. Secondly, some instances are sourced from the real data of financial firms, which is not publicly accessible online. Thirdly, the others are obtained from online financial websites. For this subset of instances, we strictly confine our selection to those published in 2023. 

\subsection{Data Statistics} All questions across CFLUE tasks undergo the standard data preprocessing pipeline, which includes processes like deduplication and the exclusion of questions relying on non-textual information. Table~\ref{tab:statistics} shows the statistics of all CFLUE tasks after preprocessing. In the knowledge assessment's multiple-choice question answering, questions are randomly divided into training, validation, and test sets at a ratio of 8:1:1, with the validation set designated for hyperparameter tuning.\footnote{We acknowledge that, in the context of few-shot and chain-of-thought evaluation, the questions in the knowledge assessment could be partitioned into dev, valid, and test sets, such as at a ratio of 1:1:8. However, in this paper, we opt not to adhere to this division due to budget constraints on OpenAI API calls.} As for tasks in the application assessment, they exclusively comprise test questions, ranging from 2.7K to 4.0K. For detailed statistics on these datasets, please refer to Table~\ref{tab:detailed_knowledge} and Figure~\ref{fig:detailed_application} in Appendix~\ref{sec:appendix_statistics}.

\section{Experimentation}

\begin{table*}[ht]
\centering
\small 
\begin{tabular}{l|lllll}
\hline
\bf Domain & \bf Model & \bf Creator & \bf \#Param. & \bf Oriented & \bf Access  \\
\hline
\multirow{8}{*}{General} & GPT-4-turbo & OpenAI & N/A & English & API \\
& GPT-4 & OpenAI & N/A & English & API \\
& ChatGPT & OpenAI & N/A & English & API \\
& LLaMA2-7B/-70B & Meta & 7B/70B & English & Weights\\
& Vicuna v1.5-7B & \citet{chiang_etal_2023_vicuna} & 7B & English & Weights \\
& ChatGLM3-6B & Tsinghua & 6B & Chinese & Weights \\
& Qwen-7B/-14B/-72B & Alibaba & 7B/14B/72B & Chinese & Weights \\
& Baichuan2-7B/-13B & Baichuan & 7B/13B & Chinese & Weights \\
\hline
\multirow{3}{*}{Financial} & FinGPT V3-6B & \citet{yang_2023_ijcai_fingpt} & 6B & Chinese & Weights\\  
& DISC-FinLLM-13B & Fudan & 13B & Chinese & Weights \\
& Tongyi-Finance-14B & Alibaba & 14B & Chinese & Weights \\
\hline
\end{tabular}
\caption{LLMs benchmarked in this paper.}
\label{tab:models}
\end{table*}

\subsection{Models, Settings, Prompts, and Metrics}
\noindent\paragraph{Models.} For a comprehensive assessment of the state of LLMs in the Chinese language context, we categorize the benchmarked LLMs into two groups: LLMs of the general domain and LLMs of the financial domain, as shown in Table~\ref{tab:models}.
\begin{itemize}[leftmargin=*]
\item \textbf{LLMs of the general domain:} This group consists of models trained on extensive volume of general-purpose corpora. Included in this group are three OpenAI LLMs \textendash GPT-4-turbo, GPT-4~\cite{openai_arxiv_2023_gpt4}, and ChatGPT (GPT-3.5-turbo)~\cite{openai_2022_chatgpt} \textendash along with five families of nine lightweight LLMs, LLaMA2-7B/-70B~\cite{touvron_arxiv_2023_llama}, Vicuna v1.5-7B~\cite{chiang_etal_2023_vicuna}, ChatGLM3-6B~\cite{zeng_etal_arxiv_2022_glm}, Qwen-7B/-14B/-72B~\cite{bai_etal_2023_arxiv_qwen}, and Baichuan2-7B/-13B~\cite{baichuan_2023_baichuan2}.
\item \textbf{LLMs in the financial domain:} This group comprises several representative LLMs specifically tailored for the financial domain. FinGPT V3-6B~\cite{yang_2023_ijcai_fingpt} is initialized with ChatGLM2-6B parameters and subsequently fine-tuned using the LoRA~\cite{hu_etal_2022_iclr_lora} method on the News and Tweets sentiment analysis dataset. DISC-FinLLM-13B~\cite{chen_etal_2023_arxiv_finllm} is based upon Baichuan-Chat-13B, with all parameters fine-tuned on Chinese financial corpora. Similarly, Tongyi-Finance-14B~\cite{tongyi_2023_tongyifinance} is based on Qwen-14B and fine-tuned on Chinese financial corpora as well.
\end{itemize}

Among the aforementioned LLMs, GPT-4 turbo, GPT-4, ChatGPT, LLaMA2, and Vicuna v1.5 are English-oriented, while the remaining models are tailored for Chinese.

\noindent\paragraph{Supervised Fine-Tuning Settings.}
For knowledge assessment, we conduct additional fine-tuning of the open-source models using the training dataset. Specifically, we employ LoRA~\cite{hu_etal_2022_iclr_lora} to fine-tune LLaMA2-7B, Vicuna v1.5-7B, ChatGLM3-6B, Qwen-7B, and Baichuan2-7B, with the rank set to 8, alpha to 32, and dropout to 0.1. During the fine-tuning, a single NVIDIA A100/80G GPU is utilized. The batch size is set to 16 with a gradient accumulation step of 8.

\noindent\paragraph{Prompts.} Our evaluation of LLMs is conducted in a zero-shot setting.\footnote{This paper focuses on the zero-shot setting, given the well-defined nature of all tasks in our assessment. We leave the exploration of the few-shot setting in our future work.} In the knowledge assessment, we follow the approach of~\citet{liu_etal_2023_nips_cmexam}, wherein we perform answer prediction and reasoning simultaneously by instructing LLMs to generate both the answer and a solution explanation. Specifically, we use three similar prompts, one for each question type. In the application assessment, we use a single prompt for each subtask, maintaining consistency by employing similar prompts for subtasks within each group. We present a few prompt examples in Appendix~\ref{sec:appendix_prompt}.

\noindent\paragraph{Metrics.} In the knowledge assessment, we measure the performance of answer prediction using accuracy and weighed F1 score. Additionally, for the open-ended reasoning we report BLEU~\cite{papineni_etal_2002_acl_bleu} and ROUGE~\cite{lin_hovy_2002_naacl_rouge} scores. In the application assessment, we report accuracy for text classification, BLEU and COMET~\cite{rei_etal_2020_emnlp_comet} for machine translation, F1 score for relation extraction, ROUGE for reading comprehension and text generation.

\begin{table*}[ht]
\resizebox{\textwidth}{!}{
\begin{tabular}{l|l|ll|lllll}
\hline
\multirow{2}{*}{\bf Domain} & \multirow{2}{*}{\bf Model} & \multicolumn{2}{c|}{\bf Prediction} & \multicolumn{5}{c}{\bf Reasoning}  \\
\cline{3-4}\cline{5-9}
& & Acc (\%) & F1 (\%) & BLEU-1 & BLEU-4 & ROUGE-1 & ROUGE-2 & ROUGE-L\\
\hline
\multirow{12}{*}{General} & GPT-4-turbo & 60.61$\pm$0.21 & 60.31$\pm$0.19 & 30.66$\pm$0.22 & 10.61$\pm$0.13 & 40.28$\pm$0.20 & 17.23$\pm$0.15 & 28.62$\pm$0.19\\
& GPT-4  & \underline{60.87$\pm$0.11} & \underline{60.82$\pm$0.10} & 37.58$\pm$0.18 & 17.26$\pm$0.09 & 44.50$\pm$0.12 & 22.42$\pm$0.08 & 32.59$\pm$0.11\\
& ChatGPT & 43.35$\pm$0.60 & 42.96$\pm$0.70 & 41.67$\pm$0.76 & 20.46$\pm$0.51 & 47.37$\pm$0.19 & 25.29$\pm$0.18 & 35.41$\pm$0.13\\
& LLaMA2-7B	& 17.66$\pm$0.39 & 10.34$\pm$0.31 & 9.46$\pm$0.16 & 3.93$\pm$0.10 & 17.77$\pm$0.17 & 7.65$\pm$0.16 & 15.48$\pm$0.18\\
& LLaMA2-70B & 18.79$\pm$0.25 & 15.54$\pm$0.21 & 13.11$\pm$0.11 & 5.49$\pm$0.07 & 22.02$\pm$0.19 & 9.72$\pm$0.14 & 19.06$\pm$0.2\\
& Vicuna v1.5-7B & 31.14$\pm$0.37 & 30.92$\pm$0.35 & 29.6$\pm$0.21 & 12.92$\pm$0.16 & 40.68$\pm$0.11 & 19.32$\pm$0.11 & 34.27$\pm$0.07\\
& ChatGLM3-6B & 40.78$\pm$0.33 & 41.37$\pm$0.33 & 34.7$\pm$0.47 & 16.74$\pm$0.23 & 43.74$\pm$0.08 & 22.92$\pm$0.09 & 37.68$\pm$0.04\\
& Qwen-7B &	43.63$\pm$0.37 & 43.25$\pm$0.41 & 42.03$\pm$0.32 & 17.85$\pm$0.29 &	39.87$\pm$0.26 & 22.11$\pm$0.21 & 35.06$\pm$0.28\\
& Qwen-14B & 53.82$\pm$0.23 & 54.23$\pm$0.27 & 40.05$\pm$0.34 & 21.56$\pm$0.25 & 47.61$\pm$0.11 & 27.27$\pm$0.10 & 41.45$\pm$0.12\\
& Qwen-72B & \bf 72.8$\pm$0.23 & \bf 73.04$\pm$0.23 & 45.78$\pm$0.39 & 26.76$\pm$0.21 & \underline{50.78$\pm$0.15} & \underline{31.48$\pm$0.13} & \bf 45.28$\pm$0.15\\
& Baichuan2-7B & 32.31$\pm$0.14 & 28.77$\pm$0.19 & 21.71$\pm$1.36 & 0.17$\pm$0.08 & 7.54$\pm$0.12 & 3.23$\pm$0.09 & 6.9$\pm$0.12\\
& Baichuan2-13B & 41.5$\pm$0.29 & 40.87$\pm$0.29 & 28.64$\pm$0.57 & 14.16$\pm$0.28 & 42.04$\pm$0.06 & 22.36$\pm$0.10 & 36.51$\pm$0.05\\
\hline
\multirow{6}{*}{Financial} & FinGPT V3-6B & 34.27$\pm$0.36 & 35.88$\pm$0.22 & 28.28$\pm$0.61 & 13.55$\pm$0.23 & 39.08$\pm$0.16 & 19.92$\pm$0.13 & 33.58$\pm$0.14 \\
 & DISC-FinLLM-13B & 35.25$\pm$0.28 & 33.79$\pm$0.24 & 29.66$\pm$0.21 & 1.69$\pm$0.15 & 14.92$\pm$0.12 & 6.56$\pm$0.18 & 11.20$\pm$0.13 \\
 & Tongyi-Finance-14B & 47.21$\pm$0.11 & 47.07$\pm$0.16 & 38.32$\pm$0.11 & 19.24$\pm$0.05 & 44.35$\pm$0.07 & 23.55$\pm$0.06 & 38.1$\pm$0.10\\
\cline{2-9}
 & LLaMA2-7B-CFLUE & 27.07$\pm$0.65 & 26.93$\pm$0.65 & 36.7$\pm$1.60 & 18.56$\pm$0.22 & 43.29$\pm$0.19 & 23.72$\pm$0.16 & 38.22$\pm$0.16\\
 & Vicuna V1.5-7B-CFLUE & 29.84$\pm$0.41 & 29.25$\pm$0.43 & 47.37$\pm$1.71 & 19.27$\pm$0.24 & 48.17$\pm$0.21 & 28.21$\pm$0.17 & 38.12$\pm$0.18\\
 & ChatGLM3-6B-CFLUE & 42.43$\pm$0.24 & 41.93$\pm$0.27 & 42.45$\pm$1.39 & 19.55$\pm$0.76 & 43.06$\pm$0.30 & 24.08$\pm$0.30 & 38.17$\pm$0.29\\ 
 & Qwen-7B-CFLUE & 48.61$\pm$0.58 & 48.59$\pm$0.6 & \bf 48.19$\pm$1.94 & \bf 22.94$\pm$1.14 & 47.62$\pm$0.19 & 27.73$\pm$0.17 & 42.41$\pm$0.15\\
 & Baichuan2-7B-CFLUE & 47.33$\pm$0.34 & 47.20$\pm$0.38 & \underline{47.69$\pm$1.45} & \underline{23.11$\pm$0.91} & \bf 53.02$\pm$0.41 & \bf 33.19$\pm$0.39 & \underline{42.98$\pm$0.27}\\
\hline
Random & Random & 21.92$\pm$0.46 & 21.58$\pm$0.46 & - & - & - & - & - \\
\hline
\end{tabular}}
\caption{Overall comparison on the answer prediction and reasoning tasks of knowledge assessment. Scores in \textbf{bold}/\underline{underline} denote the top/second-best performances.}
\label{tab:result_knowledge}
\end{table*}

\subsection{Results of Knowledge Assessment}
Table~\ref{tab:result_knowledge} compares the performance of general domain LLMs and financial domain LLMs in the contexts of answer prediction and reasoning.\footnote{Table~\ref{tab:detailed_subject} and Table~\ref{tab:detailed_question_type} in Appendix~\ref{sec:appendix_result} offer a comprehensive breakdown of performance, detailing results for each subject and question type.} Regarding the answer prediction task, we have the following observations:
\begin{itemize}[leftmargin=*]
\item GPT-4-turbo and GPT-4 exhibit superior performance compared to ChatGPT and other LLMs. These two models achieve a notable 60\% accuracy and 0.60 in F1, highlighting the challenges posed by CFLUE. Consistent with observations in other Chinese benchmarks~\cite{huang_etal_2023_nips_ceval,liu_etal_2023_nips_cmexam}, GPT-4 substantially outperform ChatGPT with a performance gap of 17\% difference in accuracy.
\item Among the nine lightweight general domain LLMs with sizes ranging from 6B to 72B parameters, Qwen-72B attains the best performance, followed by Qwen-14B and Qwen-7B which even surpass ChatGPT. The performance gap between the three Qwen models also suggests that increasing LLM size significantly benefits downstream tasks. ChatGLM3-6B and Baichuan2-13B achieve similar performance, slightly underperforming ChatGPT. Beichuan2-7B achieves 36\% accuracy and 0.35 in F1, followed by Vicuna v1.5-7B, LLaMA2-70B and LLaMA2-7B\ignore{, with the latter two performing similarly}.  
\item Among the three LLMs designed for the financial domain, Tongyi-Finance-14B attains the highest performance. Despite being built upon Qwen-14B, Tongyi Finance-14B unexpectedly lags behind with a 10\% lower accuracy. In comparison to general domain LLMs, financial domain LLMs struggle to achieve satisfactory performance. This finding aligns with that of \citet{liu_etal_2023_nips_cmexam}. Similarly, financial domain LLMs may suffer from limited corpus diversity, hindering the acquisition of broad financial knowledge needed for CFLUE questions. 
\item Fine-tuning general domain LLMs with LoRA on the CFLUE dataset significantly boosts their performance. For example, the accuracy of Qwen-7B increases to 49.84\% from 45.70\%. Even with only 4\% of the parameters used in ChatGPT, Qwen-7B-CFLUE, Baichuan2-7B-CFLUE, and ChatGLM3-6B-CFLUE all surpass ChatGPT. 
\end{itemize}

\begin{table*}[!t]
\centering
\resizebox{\textwidth}{!}{
\begin{tabular}{l|l|l|ll|ll|l|l|l|l}
\hline
\multirow{3}{*}{\textbf{Domain}} & \multirow{3}{*}{\textbf{Model}} & \multirow{3}{*}{\textbf{TC}} & \multicolumn{4}{c|}{\textbf{MT}} & \multirow{2}{*}{\textbf{RE}} & \multicolumn{1}{c|}{\multirow{2}{*}{\textbf{RC}}} & \multicolumn{1}{c|}{\multirow{2}{*}{\textbf{TG}}} & \multirow{3}{*}{\textbf{Avg.}} \\
\cline{4-7}
& & & \multicolumn{2}{c|}{\textbf{En$\rightarrow$Zh}} & \multicolumn{2}{c|}{\textbf{Zh$\rightarrow$En}} & & & & \\
\cline{3-10}
& & \multicolumn{1}{c|}{Acc.} & \multicolumn{1}{c}{BLEU} & \multicolumn{1}{c|}{COMET} & \multicolumn{1}{c}{BLEU} & \multicolumn{1}{c|}{COMET} & \multicolumn{1}{c|}{F1} & \multicolumn{1}{c|}{R-L} & \multicolumn{1}{c|}{R-L} & \\
\hline
\multirow{11}{*}{General} 
& GPT-4-turbo & \underline{60.36±0.10} &  \bf 22.81±0.08 &  \bf 79.89±0.12 & 19.90±0.04 & 87.16±0.20 &  \bf 53.81±0.29 & 44.34±0.13 & 24.22±0.09 & \underline{49.06}  \\
& GPT-4 & \bf 61.23±0.03	& 21.92±0.03 & 78.32±0.09 & 21.05±0.02 & \underline{87.20±0.13} & \underline{53.45±0.09} & \underline{46.34±0.06} & 27.55±0.05 & \bf 49.63\\
& ChatGPT & 52.42±0.16 & 21.20±0.12 & 78.21±0.11 & 19.65±0.08 & 86.82±0.11 & 52.30±0.19 & \bf 47.43±0.11 & 26.76±0.06 & 48.10  \\
& LLaMA2-7B & 4.01±0.04 & 1.59±0.05 & 28.34±0.14 & 3.37±0.06 & 34.68±0.18 & 21.48±0.25 & 4.19±0.03  & 1.09±0.01 & 12.34\\
& LLaMA2-70B & 16.67±0.50 & 3.05±0.06 & 43.19±0.35 & 4.86±0.02 & 40.59±0.16 & 26.94±0.28 & 7.07±0.10 & 6.14±0.15 & 18.56\\
& Vicuna v1.5-7B & 22.77±0.22 & 12.68±0.14 & 56.39±0.16 & 15.76±0.07 & 79.51±0.09 & 31.62±0.05 & 42.56±0.09 & 22.64±0.01 & 35.49\\
& ChatGLM3-6B & 27.65±0.01 & 14.94±0.07 & 62.40±0.14 & 16.30±0.63 & 78.26±0.16 & 23.33±0.20 & 43.08±0.10 & 26.52±0.13 & 36.56\\
& Baichuan2-7B & 18.91±0.25 & 18.78±0.53 & 50.85±0.11 & 18.11±0.11 & 52.20±0.07 & 23.29±0.11 & 24.86±0.04 & 15.46±0.12 & 32.49\\
& Baichuan2-13B & 15.06±0.10 & 19.86±0.07 & 74.44±0.06 & 19.11±0.11 & 84.15±0.05 & 31.77±0.10 & 43.45±0.11 & 28.65±0.00 & 39.56\\
& Qwen-7B & 26.07±0.62 & 18.10±0.08 & 72.53±0.13 & 19.27±0.04 & 82.69±0.11 & 35.15±0.38 & 44.36±0.05 & 28.00±0.09 & 40.77\\
& Qwen-14B & 39.87±0.26 & 19.80±0.11 & 74.99±0.09 & \underline{22.56±0.06} & 84.81±0.11 & 36.15±0.12 & 45.20±0.09 & \underline{30.11±0.08} & 44.18\\
& Qwen-72B & 51.06±0.20 & \underline{22.08±0.07} & \underline{79.20±0.03} & \bf 23.89±0.03 & \bf 87.21±0.06 & 49.21±0.11 & 43.33±0.05 & \bf 30.52±0.02 & 48.31\\
\hline
\multirow{3}{*}{Financial} 
 & FinGPT V3-6B & 19.10±0.03 & 13.90±0.12 & 60.64±0.21 & 13.63±0.08 & 73.48±0.26 & 19.16±0.24 & 39.75±0.12 & 17.33±0.05 & 32.12\\
 & DISC-FinLLM-13B & 23.24±0.06 & 15.50±0.13 & 70.95±0.12 & 4.46±0.05 & 80.63±0.14 & 32.11±0.29 & 43.32±0.08 & 24.16±0.10 & 36.80\\
 & Tongyi-Finance-14B & 29.91±0.04 & 18.98±7.63 & 73.84±0.07 & 22.41±1.87 & 84.61±0.07 & 33.32±0.16 & 45.00±0.04 & 28.85±0.02 & 42.12\\
 \hline
\end{tabular}
}
\caption{Overall comparison on the five tasks of application assessment. TC, MT, RE, RC, and TG denote text classification, machine translation, relation extraction, reading comprehension, and text generation, respectively. Scores in \textbf{bold}/\underline{underline} highlight the top/second-best performances. When calculating the average score, we initially compute the average MT score by averaging BLEU and COMET scores, and subsequently use the averaged MT score to determine the overall average score.}
\label{tab:result_application}
\end{table*}

In the reasoning task, the performance trend differs from the answer prediction task. Qwen-72B excels in BLEU and ROUGE metrics, outperforming other models. In contrast, ChatGPT surpasses GPT-4 and GPT-4-turbo, while ChatGLM3-6B and Qwen-7B match or surpass GPT-4. However, models like LLaMA2-7B, Baichuan2-7B, and DISC-FinLLM-13B exhibit subpar performance, potentially due to generating concise explanations not aligning with predicted answers. A close examination of the generated explanations indicates that, in many instances, these explanations do not align with the predicted answers. Through fine-tuning, all models show an improvement in generating more coherent explanations, achieving $\sim$20 BLEU-4 score and $\sim$40 ROUGE-L score.

\subsection{Results of Application Assessment}

\noindent\paragraph{Overall Performance.} Table~\ref{tab:result_application} provides an overview of the performance across the five application assessment tasks. GPT-4 emerges as the top performer, followed closely by GPT-4-turbo and ChatGPT. The performance gap among these three OpenAI LLMs is minimal. Among the lightweight LLMs, Qwen-72B leads, followed by Qwen-14B, Qwen-7B, Baichuan2-13B, ChatGLM3-6B, Baichuan2-7B, Vicuna v1.5-7B, LlaMA2-70B, and LlaMA2-7B. Similar to the knowledge assessment results, the three financial LLMs lag behind their general LLMs with similar parameter sizes. 

\noindent\paragraph{Results of Text Classification.} LLMs display varied text classification performance, strongly linked to model size. For example, OpenAI LLMs outperform lightweight models by a significant margin, with a performance gap exceeding 20\% in accuracy. Additionally, a substantial improvement from 26.07\% to 51.06\% is achieved by merely increasing the parameter size from 7B to 72B for Qwen. The performance gap for 6/7B LLMs (except LLaMA2-7B) is comparatively small.

\noindent\paragraph{Results of Machine Translation.} Firstly, Qwen-7B/14B/72B performance is comparable to OpenAI LLMs in machine translation. Secondly, the performance trend for BLEU and COMET is not consistently aligned. For instance, Qwen-14B and Tongyi-Finance-14B achieve the higher BLEU scores in Chinese$\rightarrow$English translation, but their COMET scores are significantly lower than those of OpenAI LLMs. DISC-FinLLM-13B attains a low BLEU score but a reasonable COMET score due to the translation involving numerous source-side words.

\noindent\paragraph{Results of Relation Extraction.} Similar to text classification, LLMs display varied performance, with OpenAI models significantly outperforming Qwen-72B by $\sim 4$\% in F1. Unlike text classification, there is no observed improvement by simply increasing the parameter size, as ChatGPT slightly underperforms GPT-4/GPT-4-turbo, while Qwen-7B performs similarly to Qwen-14B.  

\noindent\paragraph{Results of Reading Comprehension.} With the exception of LLaMA2-7B/-70B and FinGPT V3-6B, the performance gap among other LLMs is minimal, ranging from 42.56 to 47.43 in ROUGE-L. In contrast to other tasks, ChatGPT slightly outperforms GPT-4/GPT-4-turbo, while Qwen-7B performs similarly to Qwen-14B/-72B. We report BERTScore~\cite{zhang_etal_2020_iclr_bertscore} in Table~\ref{tab:rc_tg_bertscore} of Appendix~\ref{sec:appendix_result}.

\noindent\paragraph{Results of Text Generation.} OpenAI LLMs do not exhibit superiority over lightweight LLMs, especially Qwen-14B, which achieves the best performance. Similar to reading comprehension, the performance gap among all LLMs, except LLaMA2-7B/-70B, Baichuan2-7B, and FinGPT V3-6B, is relatively small (22.64 to 30.52). Similarly, we report BERTScore in Table~\ref{tab:rc_tg_bertscore} of Appendix~\ref{sec:appendix_result}.

For more detailed results per subtask in text classification, relation extraction, and text generation, please refer to Table~\ref{tab:detailed_text_classification},~\ref{tab:detailed_relation_extraction}, and~\ref{tab:detailed_text_generation} in Appendex~\ref{sec:appendix_result}. 

\section{Conclusion}
In summary, this paper have presented CFLUE, comprehensive Chinese datasets in the financial domain designed for evaluating the performance of LLMs across diverse Natural Language Processing (NLP) tasks. Comprising over 38K multiple-choice questions and 16K instances in various generation tasks, CFLUE offers a robust benchmark for assessing LLM capabilities. The detailed results presented for several LLMs on CFLUE provide valuable insights into their performance. We hope that CFLUE will contribute to future advancements in LLMs, particularly in enhancing their proficiency of Chinese NLP tasks within the financial domain.  

\section*{Ethics} CFLUE is a composite dataset sourced from various origins. Questions and corresponding solution explanations in the knowledge assessment section are derived from online mock exams. Instances in the application assessment are a combination of data from existing shared task datasets and information from select financial firms, annotated by our professional annotators.\footnote{Prior to annotation, we performed anonymization procedures on non-public financial data obtained from financial firms.} We extend full credit to the original authors of each dataset incorporated into our CFLUE benchmark, and we have secured permissions for the inclusion of each dataset in CFLUE. The data obtained from our financial firm partners is also covered by our copyright. It is important to note that all datasets within CFLUE carry low ethical risks, with stringent measures in place to ensure the absence of any sensitive or personally identifiable information.

\section*{Limitations}
Our work has several limitations. Firstly, we primarily use BLEU, COMET, ROUGE and BERTScore metrics to evaluate the performance of solution reasoning, machine translation, reading comprehension, and text generation. While informative, these metrics may not provide a comprehensive assessment of LLM outputs. Secondly, we employ identical or similar prompts for each group of subtasks in the application assessment, potentially hindering performance compared to prompts specifically tailored for each subtask. Lastly, our focus on the zero-shot setting when preparing prompts may limit LLMs from generating outputs as effectively as in a few-shot setting.

\bibliography{anthology,custom}

\appendix

\section{Data Collection for Knowledge Assessment}
\label{sec:data_knowledge_assessment}
\paragraph{Data collection.} We carefully select 15 types of qualification exams from the internet and simulated exams from universities, presenting them in either scanned or PDF format. Using tools like pdfplumber and PaddleOCR, we have collected a total of 166,681 multiple-choice questions. This extensive collection spans diverse topics within the financial domain, guaranteeing both diversity and depth in the array of questions. 

\paragraph{Data preprocessing and filtering.} We excluded questions that relied on non-text information, such as those containing HTML tags, images, tables, and questions with keywords 图/image and 表/table. Additionally, we removed duplicate questions from the dataset. 

\paragraph{Data rewriting.} We randomly shuffle the order of all options and use GPT-4 to rewrite the questions, enhancing the diversity of the dataset.

\begin{figure*}[t]
\centering
\includegraphics[width=6.5in, trim={0cm 0cm 0cm 0cm}, clip]{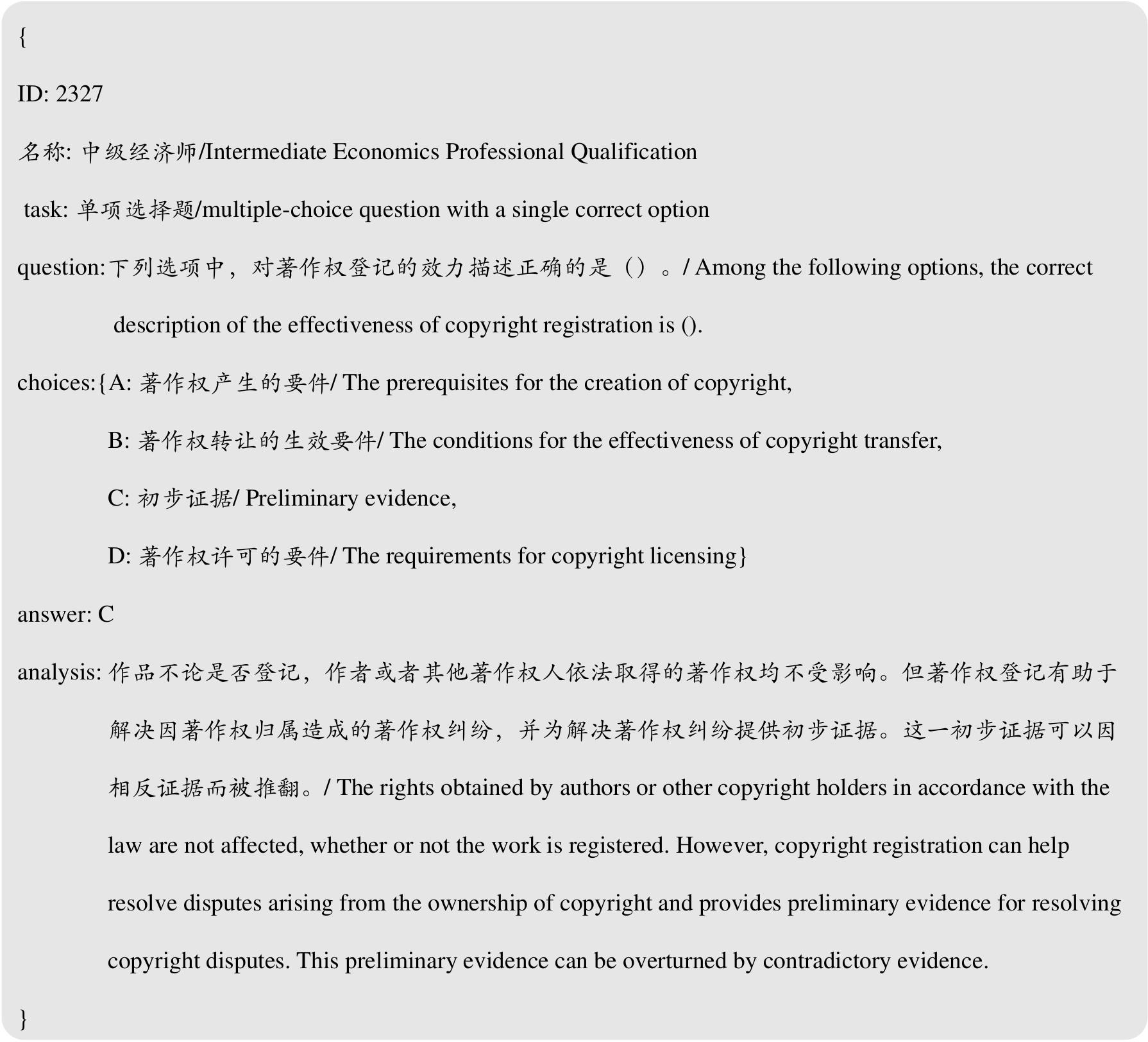}
\caption{An example of our annotated multi-choice question. English translations are provided along the corresponding Chinese text.}
\label{fig:multi-choice}
\end{figure*}

\paragraph{Data Statistics.} After preprocessing, we retain 38,636 questions. Among them, 60\% of the questions have up to six options with only one correct answer, 10\% are true/false questions, and the remaining questions may have multiple correct options. Each question in the dataset has a unique ID and is accompanied by detailed explanations provided by professionals to facilitate a thorough evaluation of the language model's logical reasoning abilities. Figure~\ref{fig:multi-choice} gives an example of annotated multi-choice question.

\ignore{
\begin{figure*}[t]
\centering
\includegraphics[width=6.5in, trim={0cm 0cm 0cm 0cm}, clip]{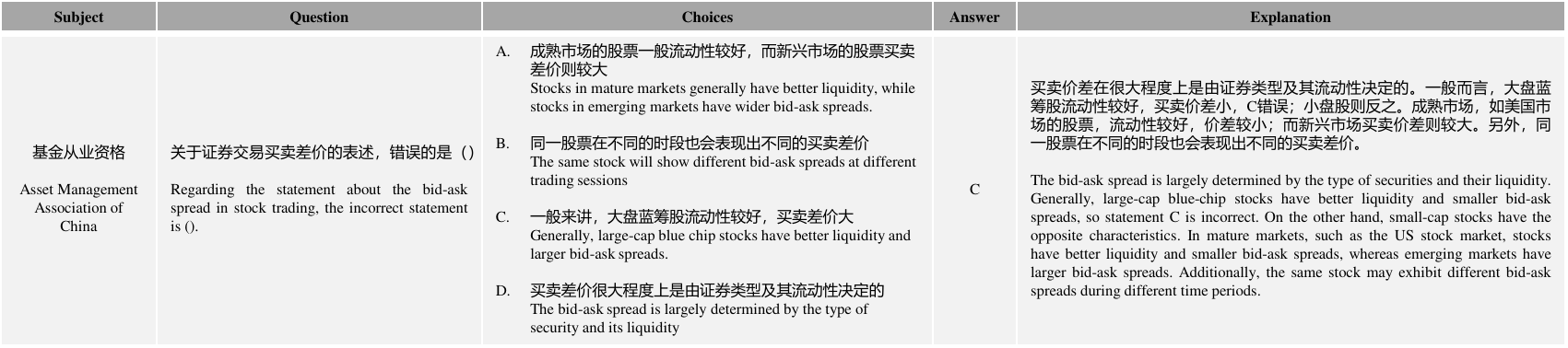}
\caption{An example question of CFLUE.}
\label{fig:example}
\end{figure*}
}

\section{Data Collection for Application Assessment}
\label{sec:data_application_assessment}

Next we detail data collection per NLP task in the application assessment in Section~\ref{sec:tc}$\sim$~\ref{sec:tg}. Then we discuss our strategy for quality control in Section~\ref{sec:quality_control}. 
\subsection{Text Classification} 
\label{sec:tc}
It covers the following five subtasks. 
\begin{itemize}[leftmargin=*]
    \item Dialogue intent classification. We randomly sample 500 instances from Banking dataset~\cite{casanueva_etal_convai_2020}\footnote{\url{https://github.com/PolyAI-LDN/task-specific-datasets}} which consists of 13,083 instances belonging to 77 intends. The sampled 500 instances are then translated into Chinese by GPT-4, followed by manual verification.
    \item ESG (environment, social and governance) classification and ESG sentiment classification. It contains 180 instances belonging to 14 classes for ESG classification and 3 classes for ESG sentiment classification. All instances are sourced from news portal websites such as 证券时报/Securities Times, 每日经济新闻/Daily Economic News, and 上海证券报/Shanghai Securities News, etc. Two senior doctoral students from \textit{anonymized} University School of Economics carry out the annotations according to the CSMAR guideline available in \url{https://data.csmar.com/}. 
    \ignore{\item Financial Document Headline Classification. It contains 54 instances belonging to 5 classes. All the instances are news articles extracted from Sina Finance\footnote{https://finance.sina.com.cn/} and categorized by the website as 规章制度/Rules and Regulations, 法律法规/Laws and Regulations, 公司公告/Company Announcements, 研究报告/Research Reports, and 财经新闻/Financial News.}
    \item Financial Event Classification. We randomly select 1,000 instances from CCKS 2020.\footnote{\url{https://tianchi.aliyun.com/dataset/111209}} These sampled instances are distributed across 27 classes, and each class has a comparable number of instances.
    \item Financial Industry Classification. It contains 1,000 instances belonging to 68 industries. All the instances are news from EastMoney website\footnote{\url{https://www.eastmoney.com/}} and are associated with the industry labeled by the website.
    \item Financial Meeting Content Classification. It contains 452 instances distributed among three classes. The ASR toolkit\footnote{\url{https://github.com/alibaba-damo-academy/FunASR}} is first utilized to convert audio files from brokerage firms' meetings into text. Annotators then divide each meeting text into various segments and categorize them as either 财务情况/Financial Situation, 未来展望/Future Outlook, or 经营情况/Operational Status. Segments that do not fall into any of these three classes are excluded.
\end{itemize}

\subsection{Machine Translation}
\label{sec:mt}
We gather Chinese-English bilingual economic news reports from \url{https://www.kekenet.com/course/13650/}. A total of 1,500 sentence pairs, each containing more than 30 words on each side, are selected. These sentence pairs serve as the dataset for both Chinese-to-English and English-to-Chinese translation.

\subsection{Relation Extraction}
\label{sec:re}
It covers the following four subtasks. 
\begin{itemize}[leftmargin=*]
\item Feature extraction in Sentiment Analysis. We collect 500 stock comments from a financial company, and annotators mark the subject of the comments along with their sentiment, covering positive, neutral, and negative. 
\item Financial Event Causality Extraction. We randomly select 1,000 instances from CCKS 2021 task6-2.\footnote{\url{https://www.biendata.xyz/competition/ccks_2021_task6_2/}}
\item Financial Event Entity Extraction. We randomly select 1,000 instances from CCKS 2019.\footnote{\url{https://www.biendata.xyz/competition/ccks_2019_4}}
\item Financial Event Extraction. We randomly select 1,000 instances from DuEE-fin~\cite{han_etal_2022_nlpcc_duee}.\footnote{\url{https://www.luge.ai/\#/luge/dataDetail?id=7}}
\end{itemize}

\subsection{Reading Comprehension}
\label{sec:rc}
Initially, we set up a document repository comprising approximately 90,000 items, which includes news articles, research reports, and annual reports. Subsequently, we generate a list of questions by drawing from the common queries encountered in real customer scenarios, involving input from over 20 financial analysts. For each question, we employ a standard extraction framework\footnote{\url{https://www.langchain.com/}} (utilizing es+m3e embedding\footnote{\url{https://huggingface.co/moka-ai/m3e-base}}) to establish a RAG (Retrieval-Augmented Generation) process, resulting in the most relevant chunk referred to as the financial text. Following this, for each (financial text, question) pair, annotators evaluate and determine if the question can be answered based on the financial text. If it is feasible, analysts write down the answer based on the financial text. Conversely, if not, analysts mark the answer as 根据提供的背景信息，无法回答该问题/According to the provided information, I cannot answer that question. After removing similar questions, we retain 2,710 triples of (financial text, question, answer), with 245 questions (9\%) that cannot be answered based on the financial text.
\subsection{Text generation}
\label{sec:tg}
It covers the following five subtasks. 
\begin{itemize}[leftmargin=*]
\item Dialogue Summarization. We randomly select 1000 instances from CSDS~\cite{lin_etal_emnlp_2021_csds}.\footnote{\url{https://github.com/xiaolinAndy/CSDS}} CSDS offers both overall summaries and role-oriented summaries. We utilize the overall summary for dialogue summarization.
\item Meeting Summarization. We employ the identical audio files utilized in Financial Meeting Content Classification. Subsequently, we invite three annotators to annotate the summary for each meeting.
\item News Headline Generation. We use the same news articles as those utilized in Financial Industry Classification. Note that each news article is associated with a corresponding headline.
\item Research Report Headline Generation. We retrieve research reports from the EastMoney website. Likewise, each research report is associated with a corresponding headline.
\item Term Interpretation. We extract financial terminologies and their explanations from ChinaValue website.\footnote{\url{http://www.chinavalue.net/Wiki}}
\end{itemize}

\subsection{Quality Control}
\label{sec:quality_control}
As shown above, the datasets involving manual annotation are from our various projects. 47 annotators in total are recruited for CFLUE. All annotators have relevant experience and education in finance, ranging from senior doctoral students and treasury analysts to finance MBAs and junior analysts. They are all trained and given access to onboarding and guidance documentation. For each project, we have additional senior analysts who review the work of annotators and provide feedback. 

\begin{itemize}[leftmargin=*]
\item For both ESG Classification and ESG Sentiment Classification, two senior doctoral students are recruited to annotate news articles . The project spanned several weeks, and tasks were assigned gradually as annotators gained confidence and experience. Weekly, each annotator received around 20-25 news articles. Five news articles were shared between both annotators, and any discrepancies in annotations for these articles were addressed through discussion and correction. Subsequently, the annotators revisited the remaining news articles.
\item For both Financial Meeting Content Classification and Meeting Summarization, 20 analysts are recruited to break down a meeting content into various segments and then categorize and summarize each segment. When provided with a text from ASR output, annotators initially segment it based on the topic.\footnote{The segmentation is performed using the automatic segmentation result from the toolkit \url{https://www.modelscope.cn/models/iic/nlp_bert_document-segmentation_chinese-base/summary}} Only segments with well-defined boundaries are chosen for classification and summarization. Before the annotation process, annotators undergo comprehensive training, and discussions involving 10 ASR texts are conducted until a high level of agreement is achieved. The project spans several weeks due to the time-consuming nature of both text segmentation and summarization.
\item For Feature extraction in Sentiment Analysis, five analysts are recruited to label the subject of the comments and their corresponding sentiment. Before the annotation process, annotators undergo training, and discussions are held using 20 comments until a high level of agreement is achieved. From the total 1,115 instances collected in the project, 500 are randomly chosen based on their lengths.  
\item For reading comprehension, 20 analysts are recruited to collect queries and mark their answer from corresponding financial text. Before the annotation process, annotators undergo training, and discussions are held using 10 pairs of (financial text, question) until a high level of agreement is achieved. Annotators discard triples if the financial texts are not relevant to their corresponding questions. The duration of this project also extended over three months, with each annotator, on average, annotating 300 triples. Out of the total 6K triples, we keep 2,710 by removing triples having similar questions.  
\end{itemize}

\section{Detailed Statistics of CFLUE}
\label{sec:appendix_statistics}
\begin{figure*}[t]
    \centering
    \includegraphics[width=\linewidth]{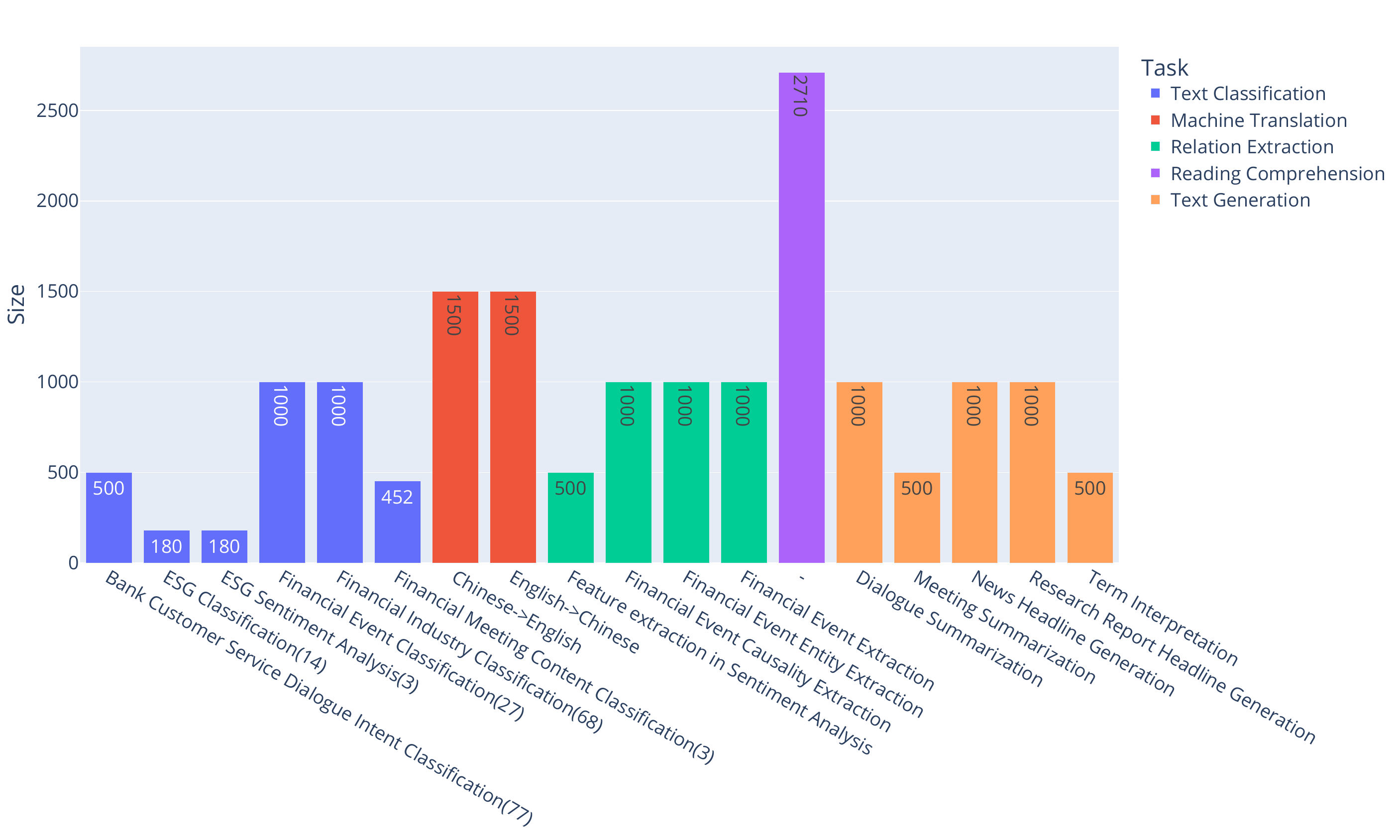}    
    \caption{Detailed statistics of the application assessment of CFLUE. For subtasks in text classification, the numbers within brackets represent the respective number of classes.} 
    \label{fig:detailed_application}
\end{figure*}

Table~\ref{tab:detailed_knowledge} lists the subjects in the knowledge assessment, as well as the number of questions included in each subject. Figure~\ref{fig:detailed_application} shows the subtasks in the application assessment, as well as the number of instances included in each subtask. 

\begin{table*}[ht]
\centering
\small
\begin{tabular}{l|lll|l}
\hline
\multirow{2}{*}{\bf Subject} & \multicolumn{4}{c}{\bf Size}\\
\cline{2-5}
& \bf Train & \bf Valid & \bf Test & \bf All\\
\hline
基金从业资格 / Asset Management Association of China & 2,691 & 336 & 337 & 3,364 \\
金融理财师 / Associate Financial Planner & 1,166 & 146 & 146 & 1,458 \\
会计从业资格 / Certificate of Accounting Professional & 2,273 & 284 & 283 & 2,840 \\
银从中级资格 / Certification of China Banking Professional (Intermediate) & 3,395 & 424 & 424 & 4,243 \\
银行初级资格 / Certification of China Banking Professional (Preliminary) & 3,617 & 453 & 452 & 4,522 \\
期货从业资格 / Certificate of Futures Qualification & 1,862 & 233 & 232 & 2,327 \\
证券从业资格 / Certification of Securities Professional & 1,223 & 153 & 154 & 1,530 \\
中国精算师 / Certified China Actuary & 103 & 13 & 13 & 129 \\
注册会计师 / Certified Public Accountant & 3,644 & 456 & 456 & 4,556 \\
保险从业资格 / China Insurance Certification \& Education & 557 & 69 & 70 & 696 \\
反假货币考试 / Counterfeit Currency Detection Exam & 486 & 60 & 61 & 607 \\
黄金从业资格 / Gold Trading Qualification Certificate & 679 & 85 & 85 & 849 \\
中级经济师 / Intermediate Economics Professional Qualification & 4,968 & 622 & 620 & 6,210 \\
初级经济师 / Junior Economics Professional Qualification & 3,279 & 409 & 411 & 4,099 \\
证券专项考试 / Securities Special Examination & 965 & 121 & 120 & 1,206 \\
\hline
Total & 30,908 & 3,864 & 3,864 & 38,636\\
\hline
\end{tabular}
\caption{Detailed statistics of the knowledge assessment of CFLUE. }
\label{tab:detailed_knowledge}
\end{table*}

\begin{figure*}[t]
    \centering
    \includegraphics[width=0.7\linewidth]{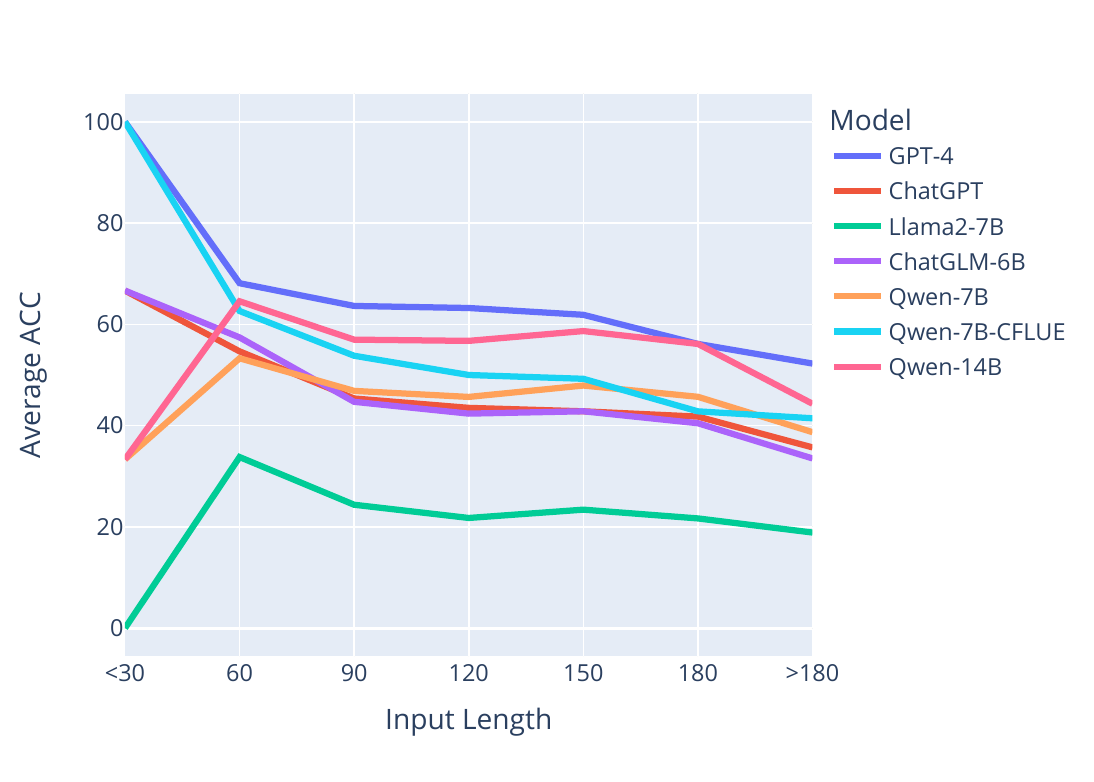}    
    \caption{Performance in accuracy over different input lengths in knowledge assessment.} 
    \label{fig:input_size_line_chart}
\end{figure*}

\section{Detailed Performance of CFLUE}
\label{sec:appendix_result}
For knowledge assessment, Table~\ref{tab:detailed_subject} and Table~\ref{tab:detailed_question_type} show the performance of answer prediction in accuracy per subject and per question type, respectively. Figure~\ref{fig:input_size_line_chart} shows the performance curves concerning input length. It shows that LLMs generally show higher accuracy with input lengths between 60 and 90. Interesting, their performance varies when the input is shorter with less then 30 tokens. 

It is important to highlight that there has been extensive research on quantitative reasoning, particularly in the context of program synthesis~\cite{kedziorski_arxiv_2023_bizbench,islam_arxiv_2023_financebench}. Our set of multiple-choice questions specifically contains quantitative reasoning questions. We utilize heuristic rules, followed by human validation, to identify quantitative reasoning questions within our test set. Consequently, we identify 334 quantitative reasoning questions, and their performance is detailed in Table~\ref{tab:detailed_quantitative_reasoning}. The results indicate that the performance on quantitative reasoning questions is significantly lower than the overall performance, suggesting the challenges that LLMs face in answering such questions. 

For application assessment, Table~\ref{tab:detailed_text_classification},~\ref{tab:detailed_relation_extraction}, and~\ref{tab:detailed_text_generation} show the performance per subtasks in text classification, relation extraction, and text generation, respectively. 

\begin{figure*}[t]
    \centering
    \includegraphics[trim=20mm 10mm 0 0, clip, width=0.7\linewidth]{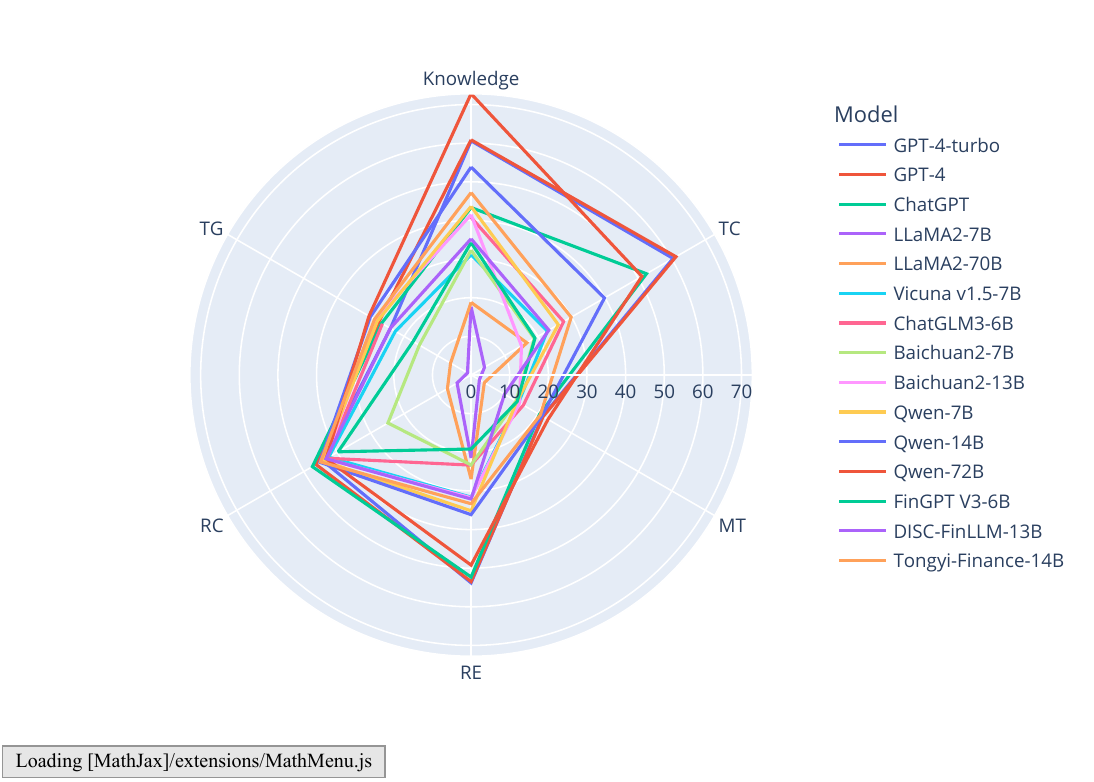}    
    \caption{Comparison among LLMs in subtasks of knowledge and application assessment.} 
    \label{fig:overall_performance}
\end{figure*}

Figure~\ref{fig:overall_performance} compares the performance of models in application assessment tasks.

\begin{table*}[ht]
\centering
\resizebox{\textwidth}{!}{
\begin{tabular}{l|lllllll}
\hline
\bf Subject & \bf GPT4 & \bf ChatGPT & \bf Qwen-7B & \bf Qwen-14B & \bf Qwen-7b-CFLUE & \bf Random\\
\hline
Asset Management Association of China & 60.87 & 71.51 & 45.7 & 55.38 & 49.84 & 26.40\\
Associate Financial Planner & 63.01 & 34.25 & 41.10 & 51.37 & 51.37 & 23.97\\
Certificate of Accounting Professional & 57.45 & 36.88 & 42.20 & 51.06 & 54.06 & 24.32\\
Certification of China Banking Professional (Intermediate) & 61.08 & 45.75 & 49.12 & 51.65 & 51.29 & 21.65 \\ 
Certification of China Banking Professional (Preliminary) & 67.04 & 52.21 & 49.12 & 61.95 & 50.66 & 25.44\\
Certificate of Futures Qualification & 57.94 & 45.92 & 42.49 & 48.93 & 52.36 & 25.49\\
Certification of Securities Professional & 53.25 & 44.81 & 49.35 & 54.55 & 54.25 & 22.59\\
Certified China Actuary & 30.77 & 53.87 & 38.46 & 38.46 & 53.85 & 23.07\\
Certified Public Accountant & 49.78 & 34.43 & 38.60 & 48.68 & 48.02 & 20.96\\
China Insurance Certification & 64.29 & 51.43 & 57.14 & 62.86 & 46.38 & 24.0\\
Counterfeit Currency Detection Exam & 57.38 & 40.98 & 42.63 & 57.38 & 55.74 & 28.52 \\
Gold Trading Qualification Certificate & 50.59 & 45.88 & 41.18 & 45.88 & 49.41 & 28.84\\
Intermediate Economics Professional Qualification & 61.67 & 40.26 & 43.16 & 52.66 & 51.04 & 15.84\\
Junior Economics Professional Qualification & 62.93 & 43.90 & 47.07 & 61.71 & 53.66 & 17.56 \\
Securities Special Examination & 69.17 & 40.83 & 47.50 & 65.00 & 51.24 & 21.16\\
\hline
\end{tabular}
}
\caption{Performance of answer prediction in accuracy per subject. To save space, we list the performance of five LLMs.}
\label{tab:detailed_subject}
\end{table*}

\begin{table*}[ht]
\centering
\small
\begin{tabular}{l|lllllll}
\hline
\bf Question Type & \bf GPT4 & \bf ChatGPT & \bf Qwen-7B & \bf Qwen-14B & \bf Qwen-7b-CFLUE & \bf Random\\
\hline
单项选择题 / with single answer & 67.70 & 49.36 & 54.01 & 64.89 & 51.98 & 25.19\\
多项选择题 / with multiple answers & 44.83 & 27.58 & 27.07 & 34.50 & 50.26 & 6.79\\
判断题 / of true or false & 68.77 & 55.87 & 52.44 & 61.60 & 55.01 & 50.60\\
\hline
\end{tabular}
\caption{Performance of answer prediction in accuracy per question type. To save space, we list the performance of five LLMs.}
\label{tab:detailed_question_type}
\end{table*}		

\begin{table*}[ht]
\centering
\small
\begin{tabular}{l|llllll}
\hline
\bf Type & \bf GPT4 & \bf ChatGPT & \bf Qwen-7B & \bf Qwen-14B & \bf Qwen-7b-CFLUE\\
\hline
quantitative reasoning & 46.41 & 26.95 & 32.63 & 32.04 & 34.13\\
\hline
\end{tabular}
\caption{Performance of answer prediction in accuracy for quantitative reasoning questions. To save space, we list the performance of five LLMs.}
\label{tab:detailed_quantitative_reasoning}
\end{table*}

\begin{table*}[ht]
\centering
\small
\begin{tabular}{l|l|l|l}
\hline
\textbf{Domain} & \textbf{Model} & \bf Reading Comprehension & \bf Text Generation \\
\hline
\multirow{11}{*}{General} 
& GPT-4-turbo & 74.20 & 69.28 \\
& GPT-4 & 75.49 & 70.01  \\
& ChatGPT & \bf 75.82 & 69.26  \\
& LLaMA2-7B & 47.84 & 45.96  \\
& LLaMA2-70B & 50.19 & 50.11  \\
& Vicuna v1.5-7B & 68.98 & 63.88  \\
& ChatGLM3-6B & 73.54 & 68.55  \\
& Baichuan2-7B & 62.37 & 58.29  \\
& Baichuan2-13B & 73.22 & 70.02  \\
& Qwen-7B & 74.25 & 69.62  \\
& Qwen-14B & \underline{74.79} & \underline{70.84}  \\
& Qwen-72B & 73.92 & \bf 71.22  \\
\hline
\multirow{3}{*}{Financial} 
 & FinGPT V3-6B & 71.30 & 48.71  \\
 & DISC-FinLLM-13B & 73.86 & 66.66  \\
 & Tongyi-Finance-14B & 74.57 & 70.06  \\
 \hline
\end{tabular}
\caption{Performance in BERTScore for reading comprehension and text generation in application assessment.}
\label{tab:rc_tg_bertscore}
\end{table*}

\begin{table*}[ht]
\centering
\small
\begin{tabular}{l|llllll}
\hline
\bf SubTask & \bf GPT4 & \bf ChatGPT & \bf Qwen-7B & \bf Qwen-14B &\bf Baichuan2-7B\\
\hline
Bank Customer Service Dialogue Intent Classification & 61.85 & 56.00 & 31.00 & 43.00 & 0.40\\
ESG Classification & 44.44 & 42.22 & 12.78 & 20.45 & 48.33\\
ESG Sentiment Analysis & 70.11 & 41.11 & 68.33 & 62.50 & 74.44\\
Financial Event Classification & 71.29 & 52.00 & 32.30 & 42.40 & 18.70 \\
Financial Industry Classification & 40.77 & 41.90 & 13.10 & 21.60 & 0.40 \\
Financial Meeting Content Classification & 82.22 & 74.78 & 15.49 & 54.55 & 41.59 \\
\hline
\end{tabular}
\caption{Performance in accuracy per subtask in text classification. To save space, we list the performance of five LLMs.}
\label{tab:detailed_text_classification}
\end{table*}

\begin{table*}[ht]
\centering
\small
\begin{tabular}{l|llllll}
\hline
\bf SubTask & \bf GPT4 & \bf ChatGPT & \bf Qwen-7B & \bf Qwen-14B &\bf Baichuan2-7B\\
\hline
Feature extraction in Sentiment Analysis & 45.15 & 36.42 & 14.20 & 27.59 & 17.35 \\
Financial Event Causality Extraction & 27.77 & 27.26 & 13.39 & 20.55 & 10.10 \\
Financial Event Entity Extraction & 88.09 & 89.20 & 87.60 & 85.29 & 74.90 \\ 
Financial Event Extraction & 53.62 & 48.36 & 15.93 & 5.63 & 17.01 \\
\hline
\end{tabular}
\caption{Performance in F1 score per subtask in relation extraction. To save space, we list the performance of five LLMs.}
\label{tab:detailed_relation_extraction}
\end{table*}

\begin{table*}[ht]
\centering
\small
\begin{tabular}{l|llllll}
\hline
\bf SubTask & \bf GPT4 & \bf ChatGPT & \bf Qwen-7B & \bf Qwen-14B &\bf Baichuan2-7B\\
\hline
Dialogue Summarization & 37.68 & 37.04 & 36.25 & 38.06 & 36.72 \\
Meeting Summarization & 35.29 & 33.04 & 33.57 & 34.75 & 31.52 \\
News Headline Generation & 21.95 & 21.45 & 21.66 & 22.33 & 23.81 \\ 
Research Report Headline Generation & 23.78 & 23.62 & 23.32 & 24.64 & 24.77 \\ 
Term Interpretation & 18.30 & 16.01 & 18.87 & 18.55 & 13.89 \\
\hline
\end{tabular}
\caption{Performance in ROUGE-L per subtask in text generation. To save space, we list the performance of five LLMs.}
\label{tab:detailed_text_generation}
\end{table*}

\section{Prompt Examples}
\label{sec:appendix_prompt}

\begin{figure*}[t]
\centering
\includegraphics[width=6.5in, trim={0cm 0cm 0cm 0cm}, clip]{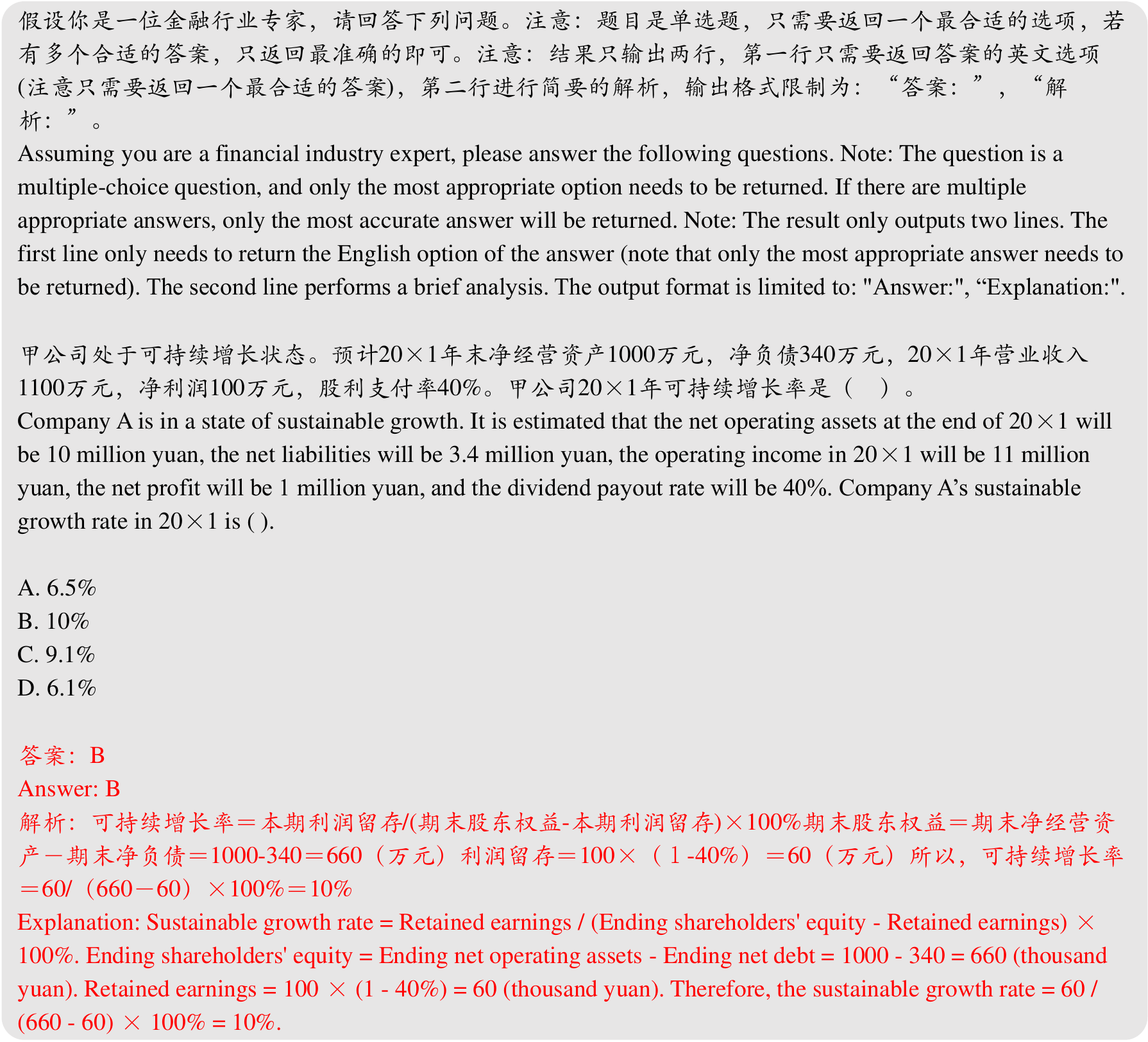}
\caption{Illustration of answer prediction and reasoning. The model generates the red text as an autocompleted response, while the preceding text serves as the input prompt. English translations are provided below the corresponding Chinese text.}
\label{fig:prompt_answer_prediction}
\end{figure*}

\begin{figure*}[t]
\centering
\includegraphics[width=6.5in, trim={0cm 0cm 0cm 0cm}, clip]{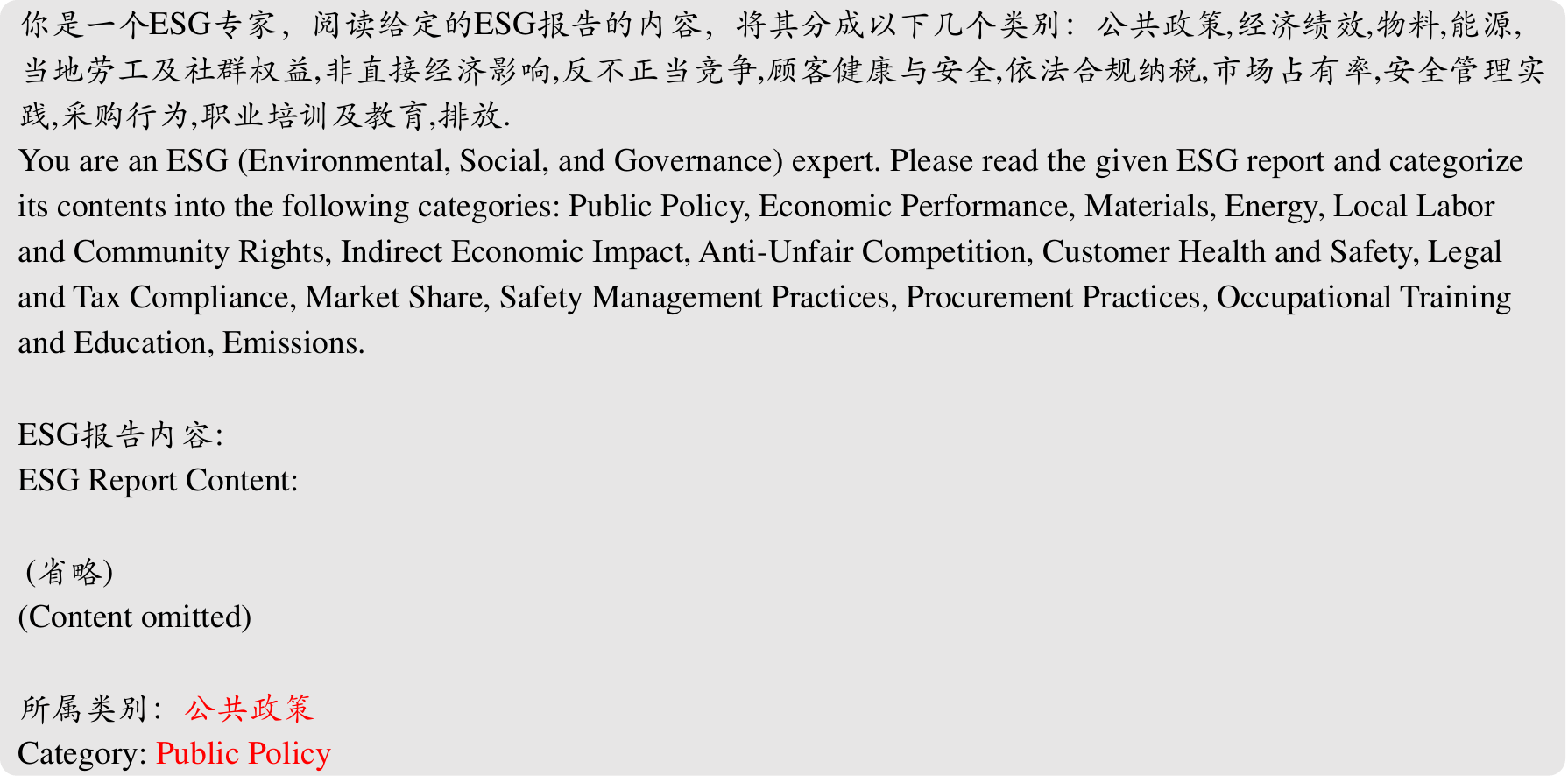}
\caption{Illustration of ESG classification subtask in text classification. The model generates the red text as an autocompleted response, while the preceding text serves as the input prompt. English translations are provided below the corresponding Chinese text.}
\label{fig:prompt_text_classification}
\end{figure*}

\begin{figure*}[t]
\centering
\includegraphics[width=6.5in, trim={0cm 0cm 0cm 0cm}, clip]{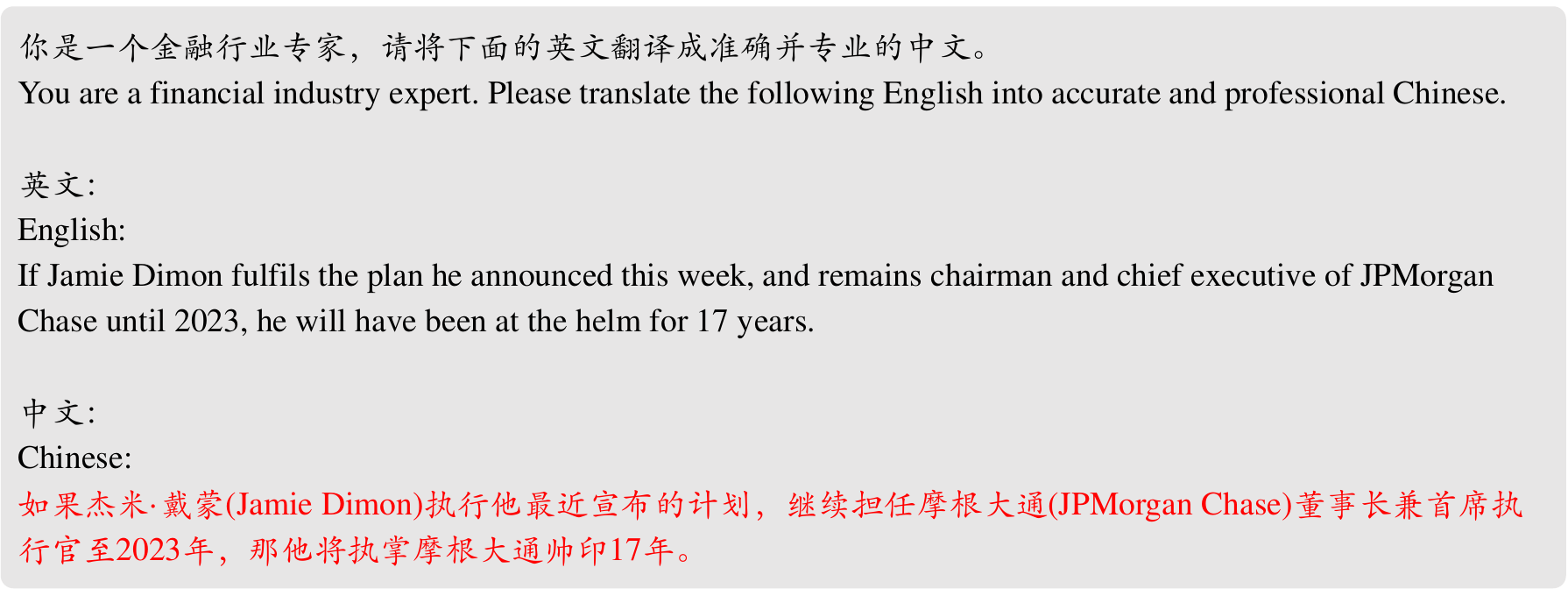}
\caption{Illustration of English$\rightarrow$Chinese subtask in machine translation. The model generates the red text as an autocompleted response, while the preceding text serves as the input prompt. English translations are provided below the corresponding Chinese text.}
\label{fig:prompt_machine translation}
\end{figure*}

\begin{figure*}[t]
\centering
\includegraphics[width=6.5in, trim={0cm 0cm 0cm 0cm}, clip]{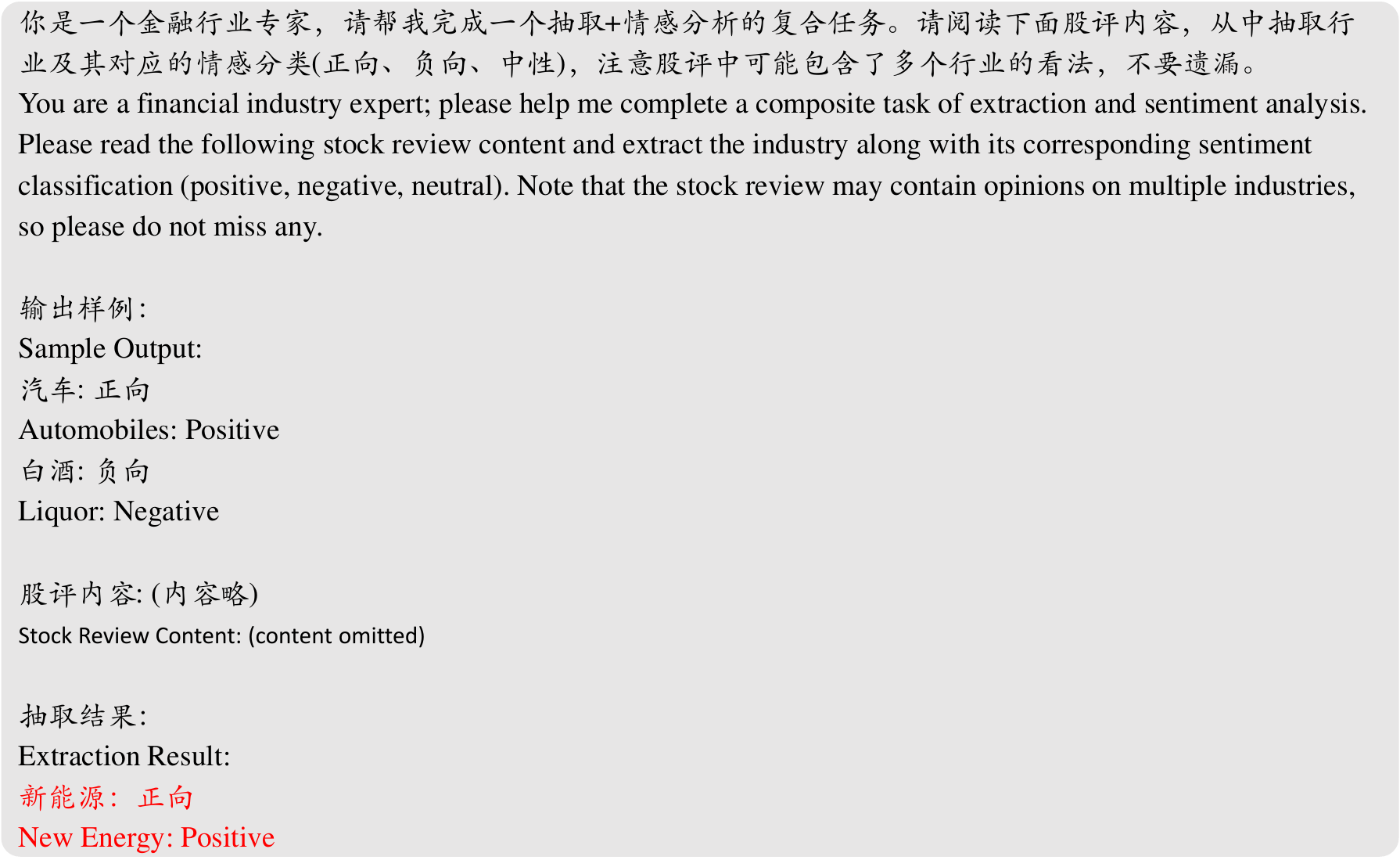}
\caption{Illustration of ESG sentiment analysis subtask in relation extraction. The model generates the red text as an autocompleted response, while the preceding text serves as the input prompt. English translations are provided below the corresponding Chinese text.}
\label{fig:prompt_relation_extraction}
\end{figure*}

\begin{figure*}[t]
\centering
\includegraphics[width=6.5in, trim={0cm 0cm 0cm 0cm}, clip]{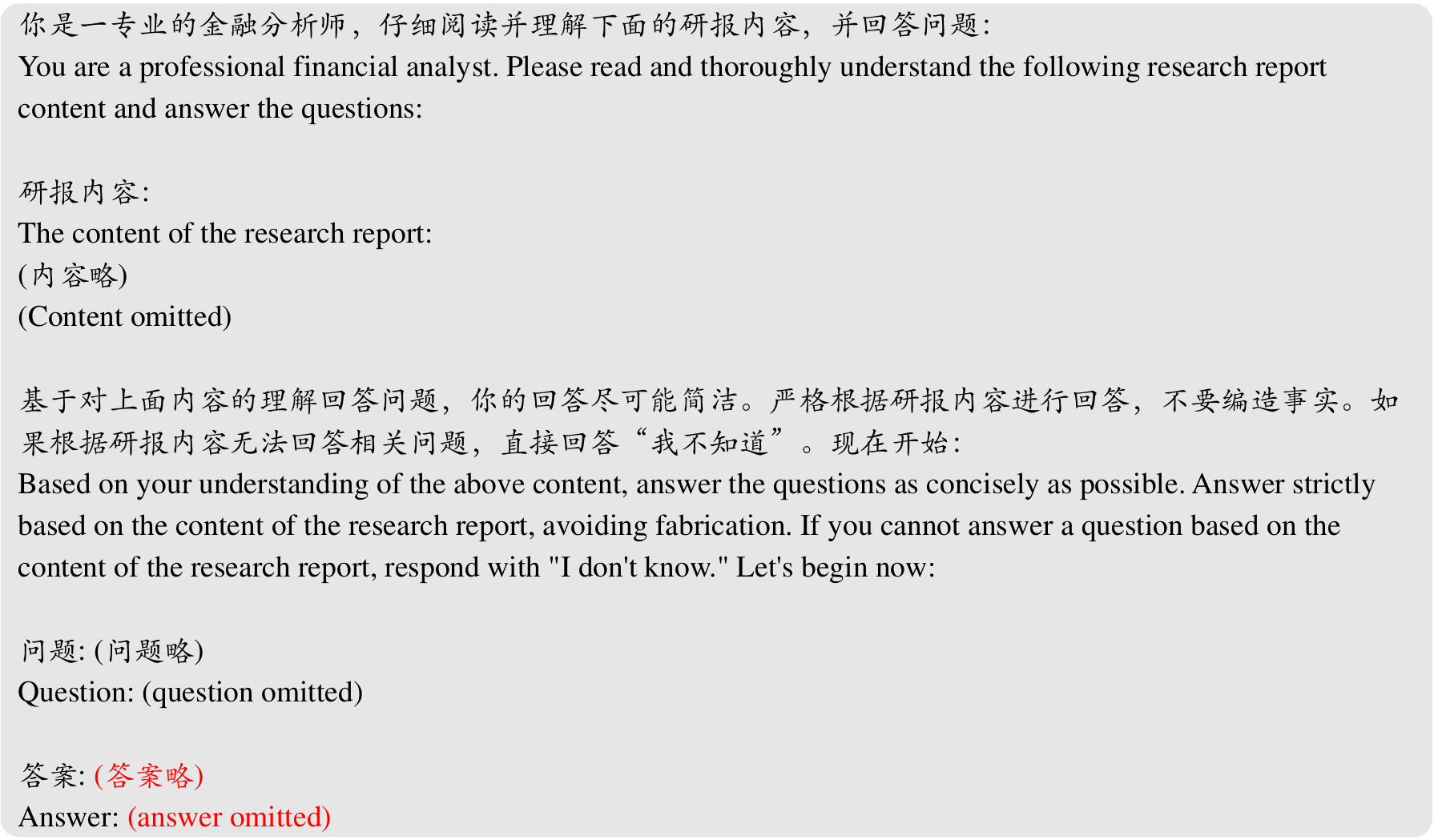}
\caption{Illustration of reading comprehension task. The model generates the red text as an autocompleted response, while the preceding text serves as the input prompt. English translations are provided below the corresponding Chinese text.}
\label{fig:prompt_reading_comprehension}
\end{figure*}

\begin{figure*}[t]
\centering
\includegraphics[width=6.5in, trim={0cm 0cm 0cm 0cm}, clip]{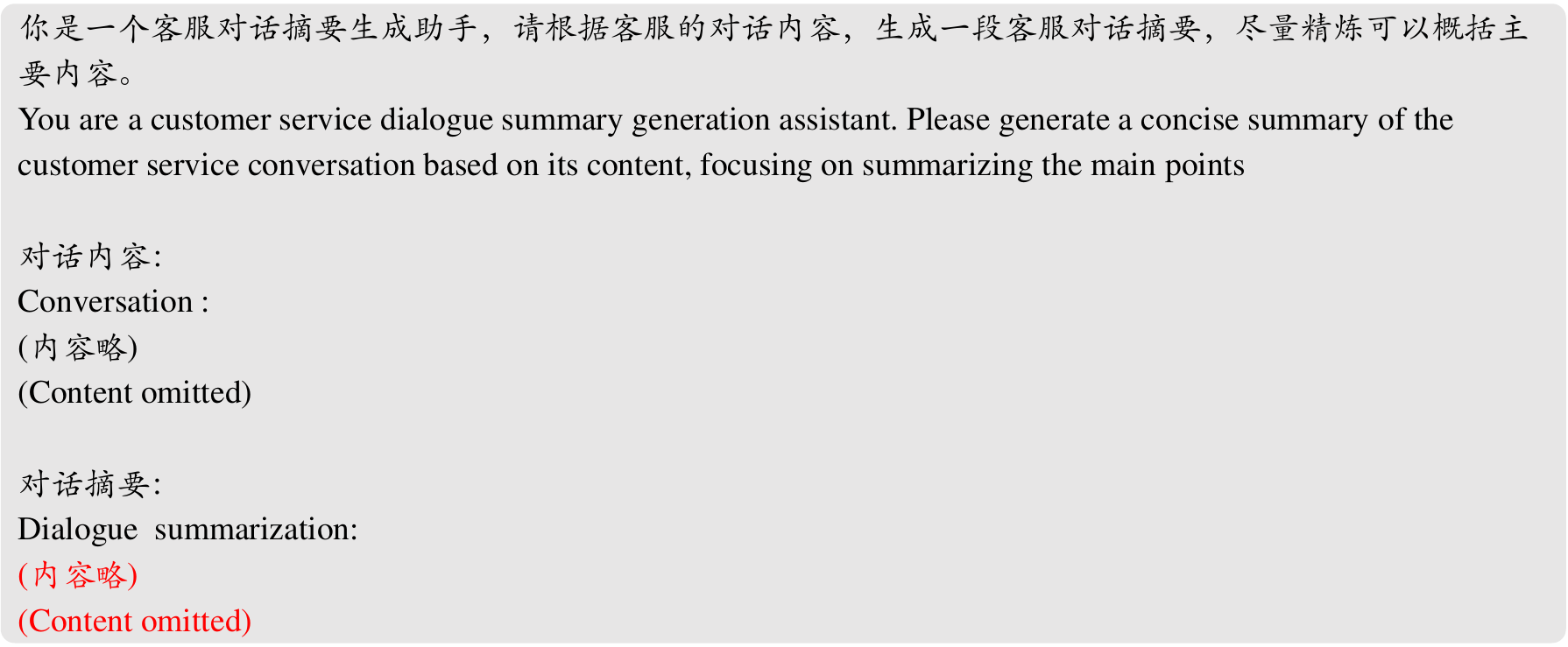}
\caption{Illustration of dialogue summarization subtask in text generation. The model generates the red text as an autocompleted response, while the preceding text serves as the input prompt. English translations are provided below the corresponding Chinese text.}
\label{fig:prompt_text_generation}
\end{figure*}

Figure~\ref{fig:prompt_answer_prediction} to~\ref{fig:prompt_text_generation} illustrate the evaluation prompts used in our experimentation.  

\end{CJK}
\end{document}